\RequirePackage{fix-cm}
\RequirePackage{fixltx2e}
\documentclass[twoside,english,aip, jcp,floatfix]{revtex4-1}
\usepackage{mathpazo}

\usepackage[T1]{fontenc}
\usepackage[latin9]{inputenc}
\usepackage[a4paper]{geometry}
\usepackage{xcolor}
\usepackage{babel}
\usepackage{mathtools}
\usepackage{amsmath}
\usepackage{amssymb}
\usepackage{graphicx}
\usepackage[unicode=true,pdfusetitle,
 bookmarks=false,
 breaklinks=true,pdfborder={0 0 0},pdfborderstyle={},backref=false,colorlinks=true]
 {hyperref}
\hypersetup{
 citecolor = blue, linkcolor = blue,  urlcolor  = blue}

\makeatletter

\providecommand{\tabularnewline}{\\}

\usepackage{orcidlink}

\makeatother

\begin{document}

\title{{\Large{}Negative impact of heavy-tailed uncertainty and }\\
{\Large{}error distributions on the reliability of calibration }\\
{\Large{}statistics for machine learning regression tasks}}

\author{Pascal PERNOT \orcidlink{0000-0001-8586-6222}}

\affiliation{Institut de Chimie Physique, UMR8000 CNRS,~\\
Université Paris-Saclay, 91405 Orsay, France}
\email{pascal.pernot@cnrs.fr}

\begin{abstract}
\noindent Average calibration of the (variance-based) prediction uncertainties
of machine learning regression tasks can be tested in two ways: one
is to estimate the calibration error (CE) as the difference between
the mean absolute error (MSE) and the mean variance (MV); the alternative
is to compare the mean squared \emph{z}-scores (ZMS) to 1. The problem
is that both approaches might lead to different conclusions, as illustrated
in this study for an ensemble of datasets from the recent machine
learning uncertainty quantification (ML-UQ) literature. It is shown
that the estimation of MV, MSE and their confidence intervals becomes
unreliable for heavy-tailed uncertainty and error distributions, which
seems to be a frequent feature of ML-UQ datasets. By contrast, the
ZMS statistic is less sensitive and offers the most reliable approach
in this context, still acknowledging that datasets with heavy-tailed
\emph{z}-scores distributions should be considered with great care.
Unfortunately, the same problem is expected to affect also \emph{conditional}
calibrations statistics, such as the popular ENCE, and very likely
\emph{post-hoc} calibration methods based on similar statistics. Several
solutions to circumvent the outlined problems are proposed.\vspace{1.5cm}
\\
\textbf{\textcolor{teal}{This preprint is a revised and extended version
of ``How to validate average calibration for machine learning regression
tasks\,?'' (https://arxiv.org/abs/2402.10043v2).}}
\end{abstract}
\maketitle

\section{Introduction}

The assessment of prediction uncertainty calibration for machine learning
(ML) regression tasks is based on two main types of statistics: (1)
the calibration errors (RCE, UCE, ENCE...)\citep{Levi2022,Pernot2023d}
which are based on the comparison of the mean squared errors (MSE)
to mean squared uncertainties or mean variance (MV); and (2) the Negative
Log-Likelihood (NLL)\citep{Gneiting2007b,Tran2020,Rasmussen2023}
which is based on the mean of squared z-scores or scaled errors (ZMS)\citep{Pernot2023d}.
The comparison of MSE to MV has been used to test or establish \emph{average}
calibration\citep{Wellendorff2014,Pernot2017}, but it mostly occurs
in ML through a bin-based setup\citep{Levi2022}, meaning that it
measures local or \emph{conditional} calibration. 

Average calibration is known to be insufficient to guarantee the reliability
of uncertainties across data space\citep{Kuleshov2018}, but it remains
a necessary condition that is too often overlooked in calibration
studies. Moreover, an interest of the RCE and ZMS statistics is to
have predefined reference values, enabling direct statistical testing
of average calibration. This is not the case for bin-based statistics
such as the UCE and ENCE, for which validation is much more complex\citep{Pernot2023a_arXiv}.
\emph{De facto}, the latter are practically used only in comparative
studies, without validation.

This study focuses on the comparison of RCE- and ZMS-based approaches
to validate calibration, and is motivated by the observation that
both approaches might lead to conflicting diagnostics when applied
to ML uncertainty quantification (ML-UQ) datasets. Understanding the
origin of such discrepancies is important to assess the reliability
of these calibration statistics and their bin-based extensions.

The next section defines the calibration statistics and the validation
approach. Sect.\,\ref{sec:ML-UQ-data} introduces the ML-UQ datasets
used to illustrate the validation results, with a focus on the shape
of their uncertainty and error distributions. Numerical experiments,
based on synthetic datasets mimicking the real datasets, are performed
in Sect.\,\ref{sec:Numerical-experiments}, in order establish the
impact of the tailedness of the uncertainty and error distributions
on the reliability of the calibration statistics. Sect.\,\ref{sec:Application}
illustrates this reliability issue on the reference ML-UQ datasets
for RCE and ZMS. The conclusions are presented in Sect.\,\ref{sec:Conclusion},
along with a proposition of solutions.
\begin{description}
\item [{\textcolor{teal}{Note}}] \textcolor{teal}{A compilation of 33 datasets
of ML materials properties published after the completion of this
study has been included }\textcolor{teal}{\emph{a posteriori}}\textcolor{teal}{{}
in Sect.\,\ref{subsec:Additional-datasets} to assess the general
validity of the main observations. }
\end{description}
\clearpage{}

\section{Average calibration statistics\label{sec:Average-calibration-statistics}}

Let us consider a dataset composed of \emph{paired} errors and uncertainties
$\left\{ E_{i},u_{E_{i}}\right\} _{i=1}^{M}$ to be tested for \emph{average}
calibration. The variance-based UQ validation statistics are built
on a probabilistic model linking errors to uncertainties
\begin{equation}
E_{i}\sim u_{E_{i}}D(0,1)\label{eq:probmod}
\end{equation}
where $D(\mu,\sigma)$ is an unspecified probability density function
with mean $\mu$ and standard deviation $\sigma$. This model states
that errors are expected to be unbiased ($\mu=0$) and that uncertainty
quantifies the \emph{dispersion} of errors, according to the metrological
definition\citep{GUM}. 

\subsection{The calibration error and related statistics\label{par:RCE}}

Let us assume that the errors are drawn from a distribution $D(0,\sigma)$
with an unknown scale parameter $\sigma$, itself distributed according
to a distribution $G$. The distribution of errors is then a \emph{scale
mixture distribution $H$, }with probability density function 
\begin{equation}
p_{H}(E)={\displaystyle \int_{0}^{\infty}p_{D}(E|\sigma)\,p_{G}(\sigma)\thinspace d\sigma}\label{eq:scale-mixture}
\end{equation}
and the variance of the compound distribution of errors is obtained
by the \emph{law of total variance}
\begin{align}
\mathrm{Var}(E) & =\left\langle \mathrm{Var}_{D}(E|\sigma)\right\rangle _{G}+\mathrm{Var}_{G}\left(\left\langle E|\sigma\right\rangle _{D}\right)\label{eq:totalVar-1}\\
 & =<u_{E}^{2}>+\mathrm{Var}_{G}\left(\left\langle E|\sigma\right\rangle _{D}\right)
\end{align}
where the first term of the RHS of Eq.\,\ref{eq:totalVar-1} has
been identified as the mean squared uncertainty $<u_{E}^{2}>$. This
expression can be compared to the standard expression for the variance
\begin{align}
\mathrm{Var}(E) & =<E^{2}>-<E>^{2}\label{eq:totalVar-1-1}
\end{align}
For an unbiased error distribution, one gets $\mathrm{Var}_{G}\left(\left\langle E|\sigma\right\rangle _{D}\right)=0$
and $<E>=0$, leading to
\begin{equation}
<E^{2}>=<u_{E}^{2}>\label{eq:RMS=00003DRMV}
\end{equation}

Based on this equation, the Relative Calibration Error, aimed to test
average calibration, is defined as
\begin{equation}
RCE=\frac{RMV-RMSE}{RMV}
\end{equation}
where $RMSE$ is the root mean squared error $\sqrt{<E^{2}>}$ and
$RMV$ is the root mean variance ($\sqrt{<u_{E}^{2}>}$). The reference
value to validate the RCE is 0. 

The RCE is rarely used as such, but it occurs in a bin-based statistic
of \emph{conditional} calibration,\citep{Pernot2023d} the Expected
Normalized Calibration Error\citep{Levi2022}
\begin{equation}
ENCE=\frac{1}{N}\sum_{i=1}^{N}|RCE_{i}|
\end{equation}
where $RCE_{i}$ is estimated over the data in bin $i$. Depending
on the variable chosen to design the bins, the ENCE might be used
to test \emph{consistency} (binning on $u_{E}$) or \emph{adaptivity}
(binning on input features)\citep{Pernot2023d}. The ENCE has no predefined
reference value (it depends on the dataset and the binning scheme)\citep{Pernot2023a_arXiv},
which complicates the statistical testing of conditional calibration.

\subsection{$ZMS$ and related statistics\label{par:ZMS}}

Another approach to calibration based on Eq.\,\ref{eq:probmod} uses
\emph{z}-scores
\begin{equation}
Z_{i}=\frac{E_{i}}{u_{E_{i}}}\sim D(0,1)
\end{equation}
with the property
\begin{equation}
Var(Z)=1
\end{equation}
assessing average calibration for unbiased errors\citep{Pernot2022a,Pernot2022b}.
If one accepts that the uncertainties have been tailored to cover
biased errors, the calibration equation becomes
\begin{equation}
ZMS=<Z^{2}>=1\label{eq:ZMS}
\end{equation}
which is the preferred form for testing\citep{Pernot2023d}, notably
when a dataset is split into subsets for the assessment of conditional
calibration. The target value for statistical validation of the ZMS
is 1. 

The negative log-likelihood (NLL) score for a normal likelihood is
linked to the ZMS by\citep{Busk2023}
\begin{align}
NLL & =\frac{1}{2}\left(<Z^{2}>+<\ln u_{E}^{2}>+\ln2\pi\right)\label{eq:NLL}
\end{align}
It combines the ZMS as an \emph{average calibration} term\citep{Zhang2023}
to a \emph{sharpness} term driving the uncertainties towards small
values\citep{Gneiting2007a} when the NLL is used as a loss function,
hence preventing the minimization of $<Z^{2}>$ by arbitrary large
uncertainties. For a given set of uncertainties, testing the NLL value
is equivalent to testing the ZMS value. 

As for the RCE, the ZMS can be used to validate conditional calibration
through a bin-based approach, the Local ZMS (LZMS) analysis\citep{Pernot2023d}.

\subsection{Validation\label{subsec:Validation-approaches-for}}

Considering that errors and uncertainties have generally non-normal
distributions, and that it is not reliable to invoke the Central Limit
Theorem to use normality-based testing approaches (see Pernot\citep{Pernot2022a}),
one has to infer confidence intervals on the statistics by bootstrapping
(BS) for comparison to their reference values. 

For a given dataset $(E,u_{E})$ and a statistic $\vartheta$, one
estimates the statistic over the dataset, $\vartheta_{est}$, and
a bootstrapped sample from which one gets the bias of the bootstrapped
distribution $b_{BS}$ and a 95\,\% confidence interval $I_{BS}=\left[I_{BS}^{-},I_{BS}^{+}\right]$.
Note that it is generally not recommended to correct $\vartheta_{est}$
from the bootstrapping bias $b_{BS}$, but it is important to check
that the bias is negligible. One of the most reliable BS approaches
in these conditions is considered to be the Bias Corrected Accelerated
(BC$_{a}$) method\citep{DiCiccio1996}, which is used throughout
this study. 

Validation is then done by checking that the target value for the
statistic, $\vartheta_{ref}$, lies within $I_{BS}$, i.e.
\begin{equation}
\vartheta_{ref}\in\left[I_{BS}^{-},I_{BS}^{+}\right]\label{eq:int-valid}
\end{equation}
To go beyond this binary result, it is interesting to have a continuous
measure of agreement, and one can define a standardized score $\zeta$
as the ratio of the signed distance of the estimated value $\vartheta_{est}$
to its reference $\vartheta_{ref}$, over the absolute value of the
distance between $\vartheta_{est}$ and the limit of the confidence
interval closest to $\vartheta_{ref}$. More concretely
\begin{equation}
\zeta(\vartheta_{est},\vartheta_{ref},I_{BS})=\begin{cases}
\frac{\vartheta_{est}-\vartheta_{ref}}{I_{BS}^{+}-\vartheta_{est}} & if\,(\vartheta_{est}-\vartheta_{ref})\le0\\
\frac{\vartheta_{est}-\vartheta_{ref}}{\vartheta_{est}-I_{BS}^{-}} & if\,(\vartheta_{est}-\vartheta_{ref})>0
\end{cases}\label{eq:zeta-def}
\end{equation}
which considers explicitly the asymmetry of $I_{BS}$ around $\vartheta_{est}$.
The compatibility of the statistic with its reference value can then
be tested by
\begin{equation}
|\zeta(\vartheta_{est},\vartheta_{ref},I_{BS})|\le1\label{eq:zeta-valid}
\end{equation}
which is strictly equivalent to the interval test (Eq.\,\ref{eq:int-valid}).
In addition to testing, $\zeta$-scores provide valuable information
about the sign and amplitude of the mismatch between the statistic
and its reference value.

\section{ML-UQ datasets\label{sec:ML-UQ-data}}

Nine test sets, including errors and \emph{calibrated} uncertainties,
have been collected from the recent ML-UQ literature for the prediction
of various physico-chemical properties by a diverse panel of ML methods.
This selection ignored small datasets and those presenting identical
properties. Note that for all these datasets, the uncertainties have
been calibrated by a palette of methods with various levels of success\citep{Pernot2023_Arxiv,Pernot2023d}.
The datasets names, sizes, bibliographic references and shape statistics
are gathered in Table\,\ref{tab:dataNu}, and the reader is referred
to the original articles for further details. In the following, a
short notation is used, e.g. 'Set 7' corresponds to the QM9\_E dataset.
\begin{table}[t]
\noindent \begin{centering}
\begin{tabular}{clcr@{\extracolsep{0pt}.}lr@{\extracolsep{0pt}.}lr@{\extracolsep{0pt}.}lr@{\extracolsep{0pt}.}l}
\hline 
{\small{}Set \# } & {\small{}Name} & {\small{}Size $M$} & \multicolumn{2}{c}{} & \multicolumn{6}{c}{Shape parameter $\nu$}\tabularnewline
\cline{6-11} 
 &  &  & \multicolumn{2}{c}{} & \multicolumn{2}{c}{ $u_{E}^{2}$} & \multicolumn{2}{c}{ $E^{2}$} & \multicolumn{2}{c}{$Z^{2}$}\tabularnewline
\cline{1-3} \cline{6-11} 
{\small{}1 } & {\small{}Diffusion\_RF\citep{Palmer2022}} & {\small{}2040 } & \multicolumn{2}{c}{} & 1&72  & 2&17  & 7&91 \tabularnewline
{\small{}2} & {\small{}Perovskite\_RF\citep{Palmer2022} } & {\small{}3834 } & \multicolumn{2}{c}{} & 0&79  & 1&18  & 4&91 \tabularnewline
{\small{}3} & {\small{}Diffusion\_LR\citep{Palmer2022} } & {\small{}2040 } & \multicolumn{2}{c}{} & 5&34  & 6&32  & 15&10 \tabularnewline
{\small{}4} & {\small{}Perovskite\_LR\citep{Palmer2022} } & {\small{}3836 } & \multicolumn{2}{c}{} & 1&53  & 2&72  & 8&15 \tabularnewline
{\small{}5} & {\small{}Diffusion\_GPR\_Bayesian\citep{Palmer2022} } & {\small{}2040 } & \multicolumn{2}{c}{} & 30&8  & 2&75  & 2&72 \tabularnewline
{\small{}6} & {\small{}Perovskite\_GPR\_Bayesian\citep{Palmer2022} } & {\small{}3818 } & \multicolumn{2}{c}{} & 1&19  & 0&78  & 0&85 \tabularnewline
{\small{}7} & {\small{}QM9\_E\citep{Busk2022}} & {\small{}13885 } & \multicolumn{2}{c}{} & 1&91  & 2&43  & 3&95 \tabularnewline
{\small{}8 } & {\small{}logP\_10k\_a\_LS-GCN\citep{Rasmussen2023} } & {\small{}5000 } & \multicolumn{2}{c}{} & 24&7  & 4&24  & 3&66 \tabularnewline
{\small{}9 } & {\small{}logP\_150k\_LS-GCN\citep{Rasmussen2023} } & {\small{}5000 } & \multicolumn{2}{c}{} & 17&3  & 10&0  & 20&2 \tabularnewline
\hline 
\end{tabular}
\par\end{centering}
\caption{\label{tab:dataNu}The nine datasets used in this study: number, name
with reference, size, and shape parameters for the fits of $u_{E}^{2}$
by an Inverse Gamma distribution, and of the $E^{2}$ and $Z^{2}$
by an \emph{F} distribution. }
\end{table}

As the shape of the distributions of $u_{E}^{2}$, $E^{2}$ and $Z^{2}$
is central to the elucidation of the problem presented in the following
sections, these were characterized for each dataset as explained below
and reported in Table\,\ref{tab:dataNu}. 

\subsection{$u_{E}^{2}$ distributions}

The Inverse-Gamma (IG) distribution\citep{Evans2000} is commonly
used to describe variance ($u_{E}^{2}$) samples
\begin{equation}
u_{E}^{2}\sim\Gamma^{-1}(\alpha=\nu,\beta=\sigma\nu)
\end{equation}
where $\alpha>0$ and $\beta>0$ are the shape and scale parameters,
respectively, expressed here as combinations of a number of degrees
of freedom $\nu$ and a scale factor $\sigma$. This equation is more
conveniently expressed as 
\begin{equation}
u_{E}^{2}/\sigma^{2}\sim\Gamma^{-1}(\nu,\nu)\label{eq:IG}
\end{equation}
This distributions is often used in Bayesian inference as a prior
for variance parameters. 
\begin{figure}[t]
\noindent \begin{centering}
\includegraphics[width=0.85\textwidth]{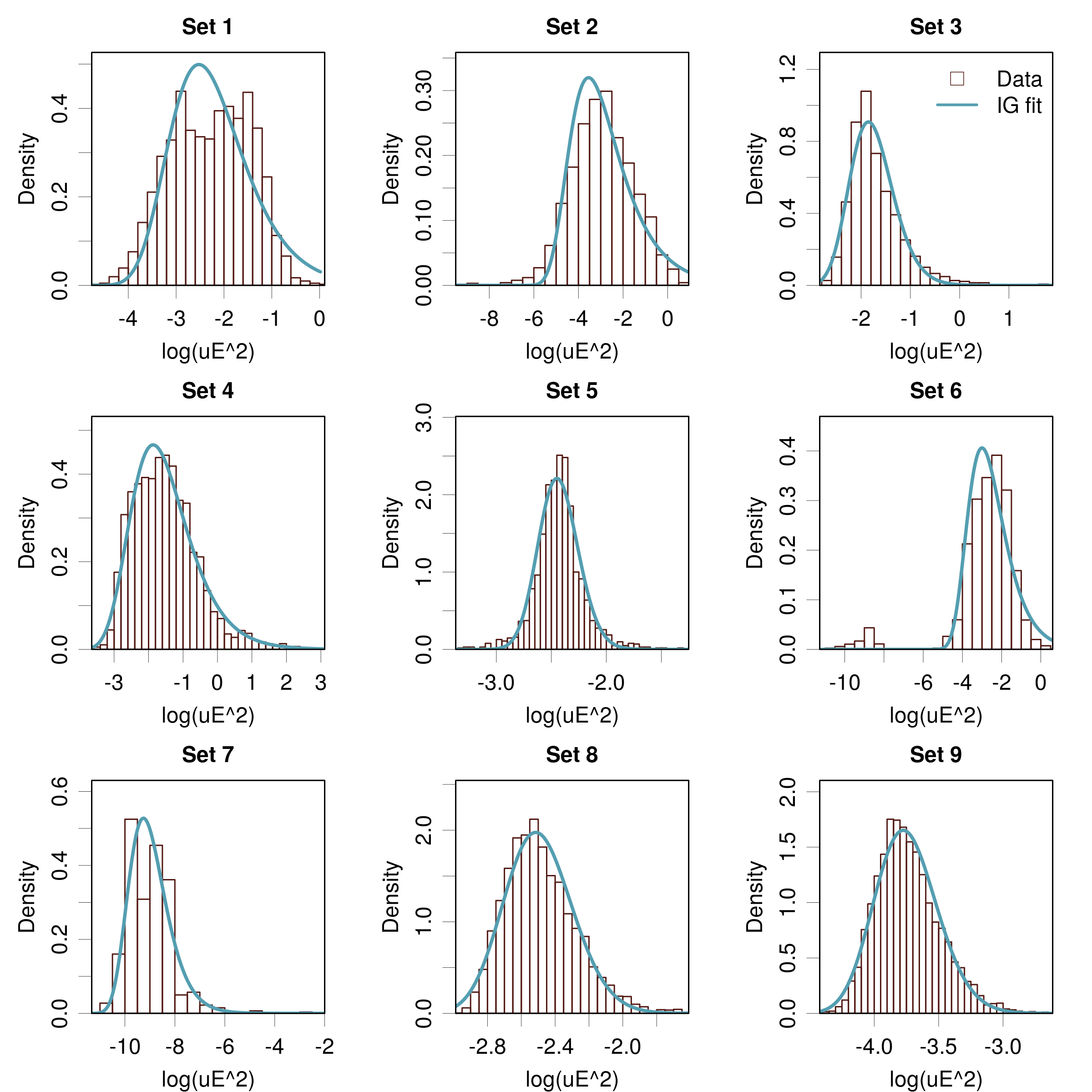}
\par\end{centering}
\caption{\label{fig:fituE2}Fit of the squared uncertainties (histogram) by
an Inverse-Gamma $\Gamma^{-1}(\nu,\nu)$ distribution (blue line). }
\end{figure}

Fit of squared uncertainty distributions by the IG model is done by
maximum goodness-of-fit estimation using the Kolmogorov-Smirnov distance
\citep{Delignette2015}. The results are shown in Fig.\,\ref{fig:fituE2},
and the shape parameters are reported in Table\,\ref{tab:dataNu}.

Overall, the fits are of contrasted quality. The distributions for
Sets 1, 2 and 6 are the worst ones, mostly due to the presence of
a bimodality (a tiny one for Set 2). The $\nu$ values for these sets
are very low: for Set 2 $\nu<1$ corresponds to an IG distributions
with undefined mean value, whereas for Sets 1 and 6 values of $\nu<2$
indicate an undefined variance. These values should not be over-interpreted.
Nevertheless, better fitted distributions present also very small
shape parameters ($\nu<2$), such as Sets 4 and 7, meaning that such
values might not be unreasonable. 

One can thus characterize the variance datasets by their shape parameter
$\nu$, which covers a wide range from 0.8 to 31 (Table\,\ref{tab:dataNu}).

\subsection{$E^{2}$ distributions}

Using the generative model, Eq.\,\ref{eq:probmod}, with a standard
normal distribution {[}$D=N(0,1)${]} and an IG distribution of squared
uncertainties $u_{E}^{2}/\sigma^{2}\sim\Gamma^{-1}(\nu/2,\nu/2)$
leads to a scale mixture distribution of errors (Eq.\,\ref{eq:scale-mixture}),
which has the shape of a Student's-$t$ distribution with $\nu$ degrees
of freedom\citep{Andrews1974,Choy2008}
\begin{equation}
E/\sigma\sim t(\nu)\label{eq:NIG}
\end{equation}
This scale mixture is a sub-case of the Normal-IG (\emph{NIG}) distribution
used in evidential inference.\citep{Amini2019} For the squared errors,
one gets 
\begin{equation}
E^{2}/\sigma^{2}\sim F(1,\nu)
\end{equation}
where $F(\nu_{1},\nu_{2})$ is the Fisher-Snedecor (\emph{F}) distribution\citep{Evans2000}
with degrees of freedom $\nu_{1}$ and $\nu_{2}$. 
\begin{figure}[t]
\noindent \begin{centering}
\includegraphics[width=0.85\textwidth]{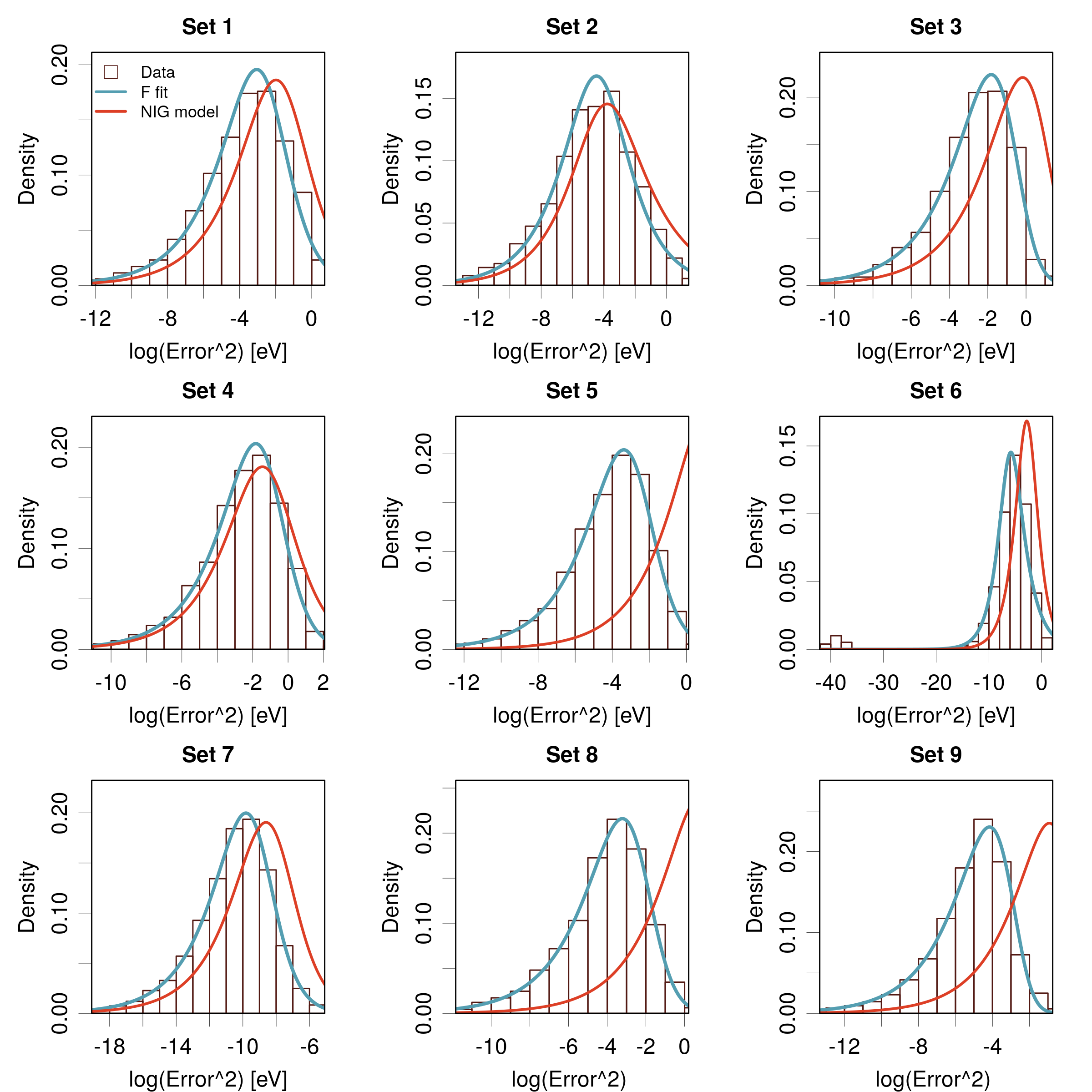}
\par\end{centering}
\caption{\label{fig:fitE2}Fit of the squared errors (histogram) by a Fisher-Snedecor
$F(1,\nu)$ distribution (blue line). The red curves represent the
distributions for NIG models compatible with Fig.\,\ref{fig:fituE2}. }
\end{figure}

The squared error datasets have been fitted by a scaled $F$ distribution,
and the results are reported in Fig.\,\ref{fig:fitE2} and Table\,\ref{tab:dataNu}.
The fits are very good for all sets, except for Set 6, where a minor
mode at small errors cannot be accounted for by the $F$ distribution.
Here again, the shape parameters cover a wide range, from 0.8 to 10,
with very values typical of heavy tails. 

It has to be noted that the shape parameter for $E^{2}$ is never
close to being twice the shape parameter for $u_{E}^{2}$ as expected
from the NIG model (Eq.\,\ref{eq:NIG}). The densities expected from
the NIG model are plotted in Fig.\,\ref{fig:fitE2} (red lines) for
comparison with the actual fits and confirm the discrepancy. Therefore,
either the generative distribution $D$ is never close to $N(0,1)$
for the studied datasets, or the datasets are not properly calibrated. 

\subsection{$Z^{2}$ distributions}

The shape parameters for the fit of $Z^{2}$ distributions by the
same procedure as for $E^{2}$ are reported in Table\,\ref{tab:dataNu}.
The fits (not shown) present the same features as $E^{2}$ fits, albeit
with typically larger shape parameters, ranging from 0.9 to 20, which
is expected from the generative model. Two exceptions are Sets 5 and
8, for which the shape parameter is slightly smaller for $Z^{2}$
than for $E^{2}$, a small difference that might not be significant.

\subsection{Summary}

This section showed that it was possible to represent the distribution
of $u_{E}^{2}$ sets by an IG distribution with reasonable accuracy,
except for bimodal distributions, and that some of these distributions
have very small shape parameters indicating very heavy tails. The
$E^{2}$ and $Z^{2}$ distributions were successfully fit by $F$
distributions, some with very small shape parameters indicating heavy
tails, but the optimal shape parameters did not conform with the constraints
of the NIG model. 

The next section demonstrates that the presence of heavy tails in
either of these quantities might challenge the reliable estimation
and validation of calibration statistics.\clearpage{}

\section{Numerical experiments\label{sec:Numerical-experiments}}

The characteristic features of the nine datasets presented in the
previous sections are now used to design numerical experiments to
assess the impact of the distributions tails on the reliability of
calibration statistics. A first step is to define tailedness metric
independent of model distributions. 

\subsection{Alternative tailedness metrics\label{subsec:Skewness-as-a}}

As shown in the previous section, the shape of the uncertainty and
error distributions can be rather well characterized by a shape parameter
$\nu$, but one has to take into account that not all datasets are
correctly fitted by the model distributions. Besides, for the $IG$
and $t$ distributions, shape metrics such as skewness and kurtosis
are not defined for $\nu\le3$ or 4, respectively. One therefore needs
distribution-agnostic metrics able to characterize the shape of these
distributions.

Considering the asymmetric shape of $u_{E}^{2}$, $E^{2}$ and $Z^{2}$
distributions, one might want to describe the length of the upper
tail by skewness and/or the heaviness of the tails by kurtosis. For
such distributions, these metrics are expected to be correlated\citep{Pernot2021},
but they might still provide complementary information.

As one is potentially dealing with heavy-tailed distributions and
outliers, it is essential to use robust statistics. $\beta_{GM}$
is a skewness statistic based on the scaled difference between the
mean and median \citep{Crow1967,Groeneveld1984,Bonato2011,Pernot2021},
which is robust to outliers, varies between -1 and 1 and is null for
symmetric distributions. For kurtosis, $\kappa_{CS}$ is chosen for
the same reasons\citep{Crow1967,Groeneveld1984,Bonato2011,Pernot2021}.
This is an excess kurtosis, meaning that positive values indicate
tails that are heavier than those of the normal distribution. $\kappa_{CS}$
is not scaled and does not have finite limits.

\begin{figure}[t]
\noindent \begin{centering}
\includegraphics[width=0.48\textwidth]{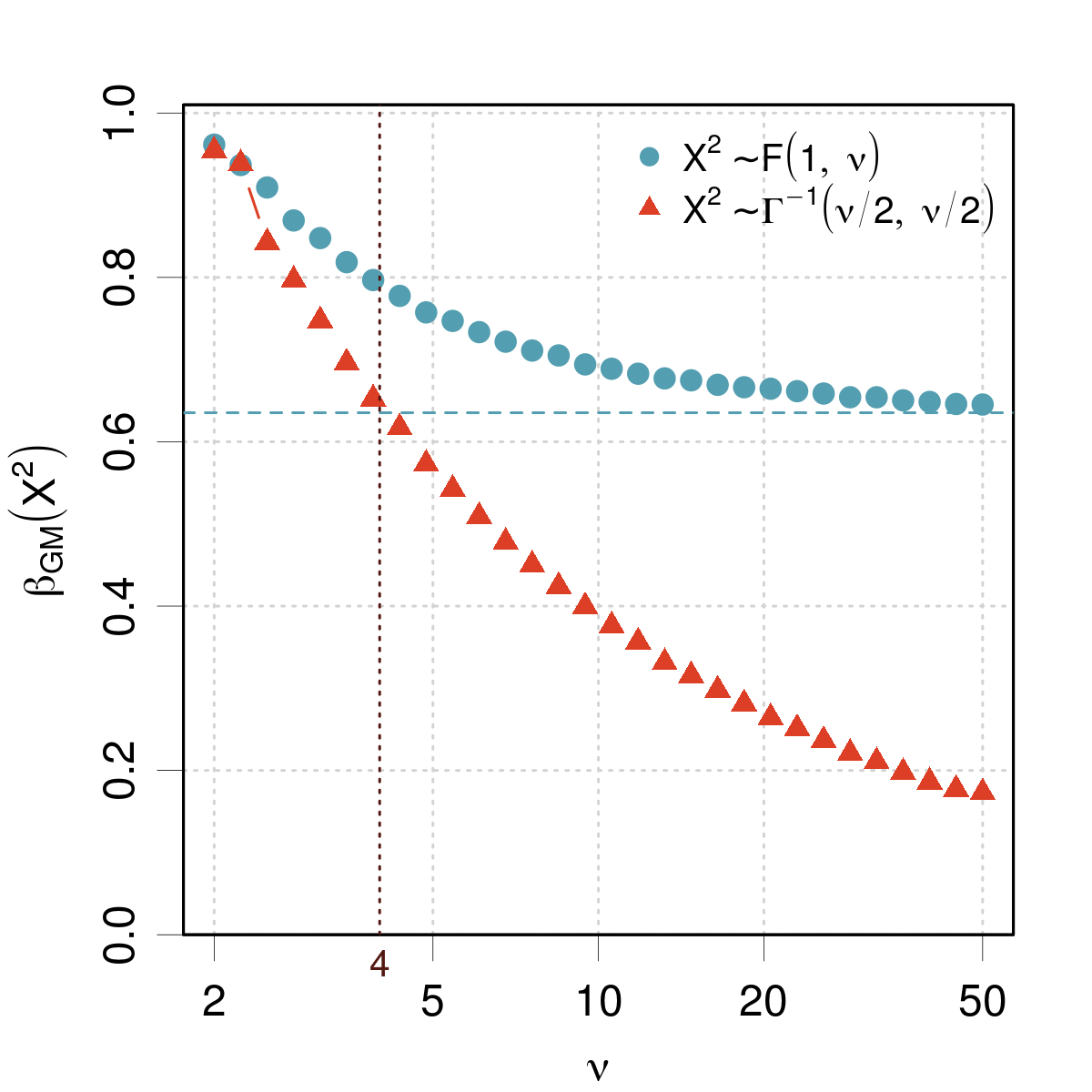}\includegraphics[width=0.48\textwidth]{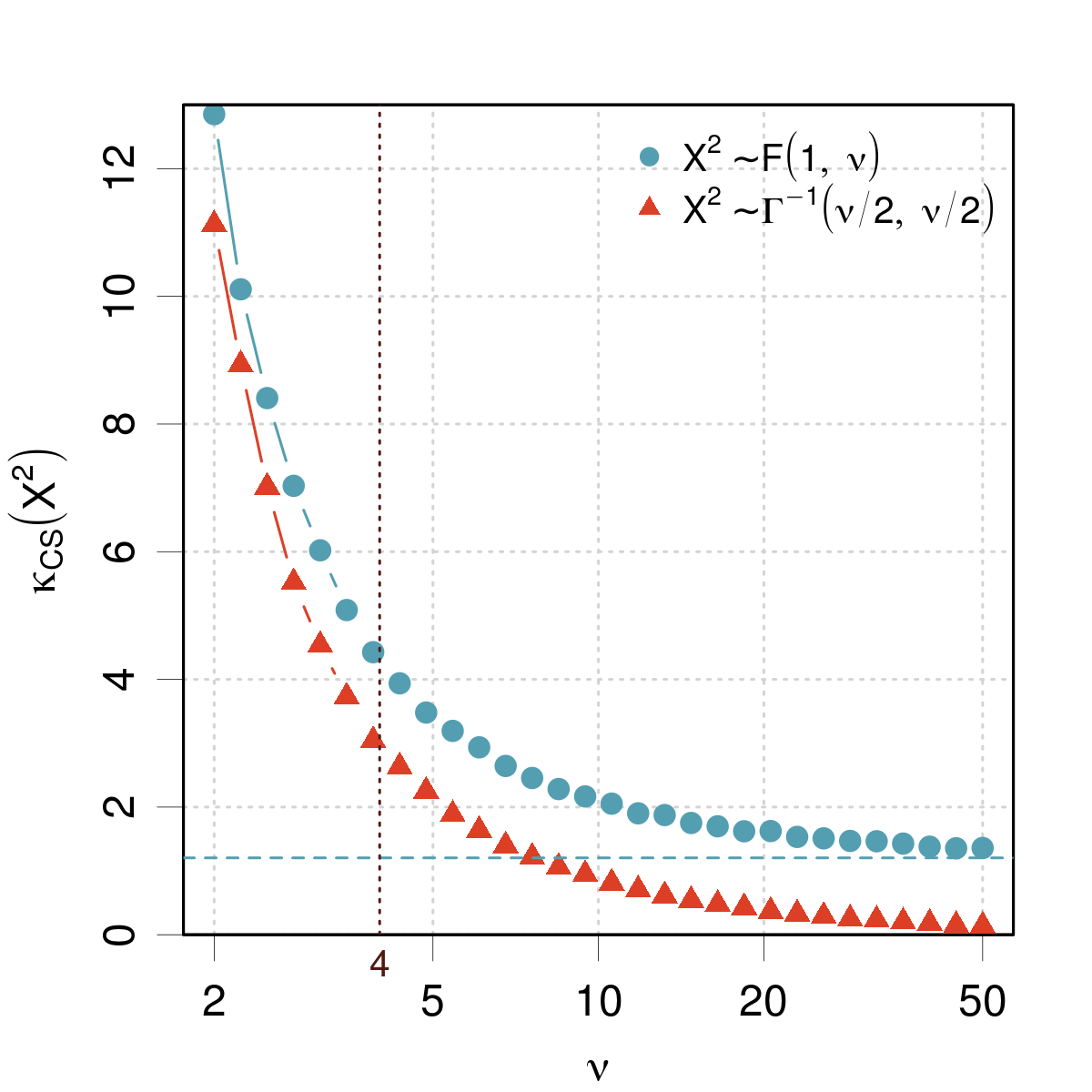}
\par\end{centering}
\caption{\label{fig:beta}$\beta_{GM}$ skewness and $\kappa_{CS}$ kurtosis
values for samples (size $5\times10^{5}$) issued from Fisher-Snedecor
$F(1,\nu)$ distributions (blue dots) and from Inverse-Gamma $\Gamma^{-1}(\nu/2,\nu/2)$
distributions (red triangles). The dashed horizontal line represents
the limit for a squared normal variate. }
\end{figure}

Fig.\,\ref{fig:beta} shows the correspondence between $\nu$, $\beta_{GM}$
and $\kappa_{CS}$ for samples of the \emph{IG} distribution {[}$X^{2}\sim\Gamma^{-1}(\nu/2,\nu/2)${]}
and \emph{F}-distribution {[}$X^{2}\sim F(1,\nu)${]}. There is a
monotonous one-to-one correspondence between $\beta_{GM}$ , $\kappa_{CS}$
and the $\nu$ parameter of the sampled distributions showing that
the skewness and kurtosis statistics can be used to replace unambiguously
the shape parameter of the model distributions. 

Note that the coefficient of variation $c_{v}$ is often used in ML-UQ
studies to quantify the dispersion of uncertainties\citep{Scalia2020},
which is also related to the tailedness of their distribution. However,
it is not robust and cannot replace the proposed statistics. 

Considering that $\beta_{GM}$ and $\kappa_{CS}$\textcolor{orange}{{}
}are practically exchangeable for the kind of distributions considered
here, only $\beta_{GM}$ is used in the following to characterize
the tailedness of the distributions. It provides a conveniently bounded
metric.

\subsection{Sensitivity of calibration statistics to heavy-tailed uncertainty
and error distributions\label{subsec:Sensitivity-of-calibration}}

\begin{figure}[t]
\noindent \begin{centering}
\includegraphics[width=0.96\textwidth]{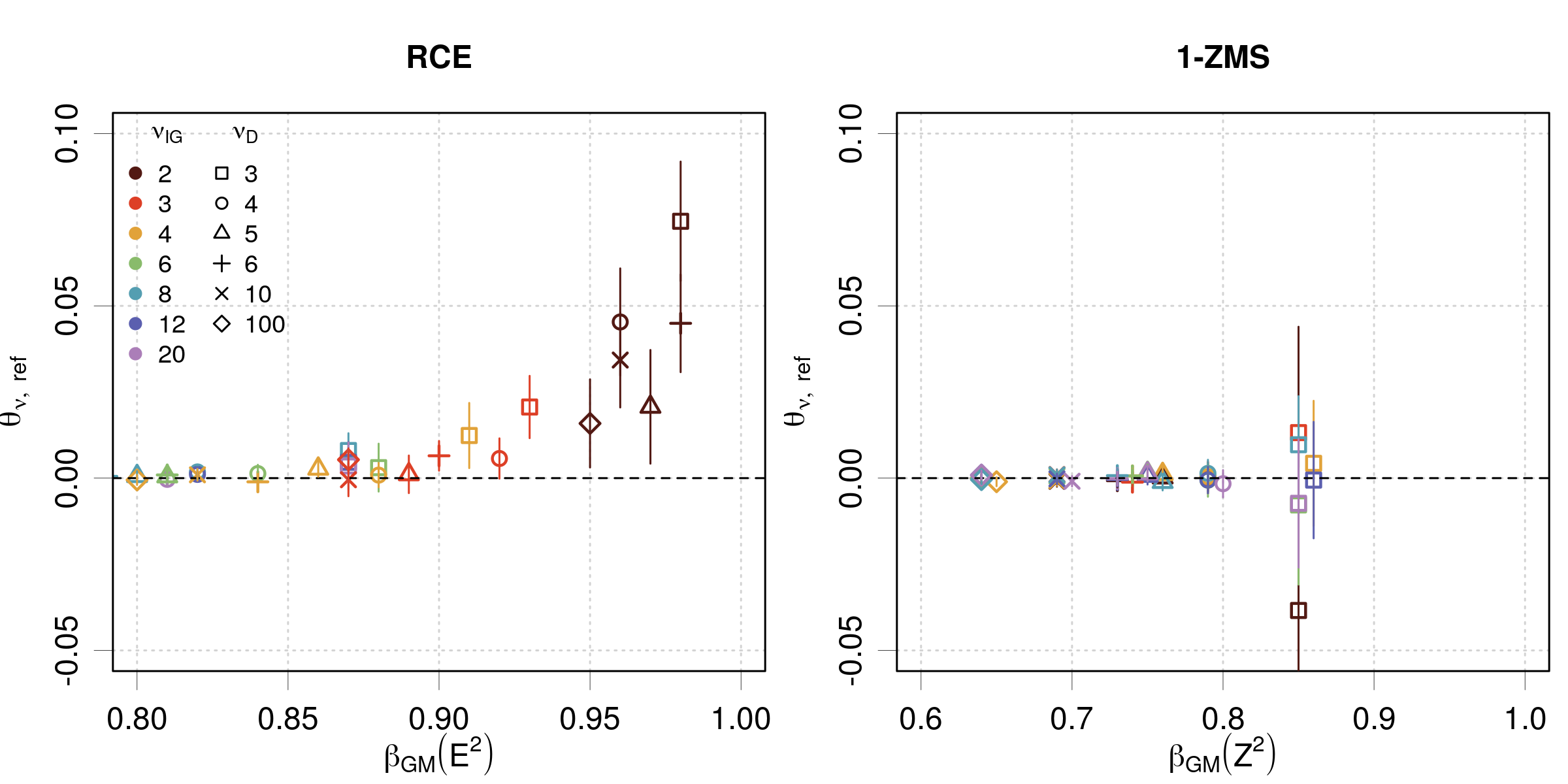} 
\par\end{centering}
\caption{\label{fig:altRCE-1}Comparison of the estimated values of RCE and
1-ZMS for a series of datasets generated by TIG models and characterized
by their error skewness parameter $\beta_{GM}(E^{2})$, or $\beta_{GM}(Z^{2})$
for 1-ZMS. The symbols and 2$\sigma$ error bars summarize a sample
of $10^{3}$ Monte Carlo runs.}
\end{figure}

To characterize the sensitivity of the RCE and ZMS statistics to the
shape of the uncertainty and error distributions, they are estimated
for synthetic datasets covering the range of tail statistics observed
in the previous sections. $N=10^{3}$ synthetic calibrated datasets
of size $M=5000$ are generated using an extension of the NIG model,
the TIG model, defined by $u_{E}^{2}\sim\Gamma^{-1}(\nu_{IG}/2,\nu_{IG}/2)$
and $D=t_{s}(\nu_{D})$) where
\begin{equation}
t_{s}(\nu_{D})\coloneqq t(\nu_{D})\times\sqrt{\frac{\nu_{D}-2}{\nu_{D}}}\label{eq:TS}
\end{equation}
for combinations of parameters such as $\nu_{IG}\in[2,20]$ and $\nu_{D}\in[3,100]$.\textcolor{orange}{{}
}Smaller values of the parameters have been avoided as they might
result in numerical problems. The results are reported in Fig.\,\ref{fig:altRCE-1}
for RCE vs. $\beta_{GM}(E^{2})$ and 1-ZMS vs. $\beta_{GM}(Z^{2})$.
The mean values and 2$\sigma$ error bars summarize the sample of
$N$ Monte Carlo runs.

The RCE deviates to positive values when $\beta_{GM}(E^{2})>0.85$,
and the deviation is increasing with skewness. As the Monte Carlo
error bars are not large enough to cover the reference value (0),
one can conclude that the RCE sis significantly biased in this area.
Note that reformulating the RCE without square roots can help to reduce
this bias (see Appendix\,\ref{sec:An-alternative-formulation}).
In comparison, the 1-ZMS statistic is better behaved for most of the
datasets, and seems to have problems only with the extreme $\nu_{D}=3$
case, which corresponds to $\beta_{GM}(Z^{2})>0.8$.

\subsection{Sensitivity of validation statistics to heavy-tailed uncertainty
and error distributions\label{sec:Compared-reliability-of}}

The previous experiment revealed how the RCE and ZMS are sensitive
to heavy tails. It occurs that the estimation of confidence intervals
for these statistics by bootstrapping has also difficulties in similar
conditions, which is put in evidence by the following simulations.

One considers here two scenarios designed to cover the range of distributions
observed above, but also to limit the computational cost of the experiments:
\begin{itemize}
\item NIG: $u_{E}^{2}\sim\Gamma^{-1}(\nu_{IG}/2,\nu_{IG}/2)$ and $D=N(0,1)$,
with $\nu_{IG}$ varying between 2 and 10.
\item TIG: $u_{E}^{2}\sim\Gamma^{-1}(3,3)$ and $D=t_{s}(\nu_{D})$ with
$\nu_{D}$ varying between 2.1 and 20. 
\end{itemize}
For each sample ($M=5000$ with $N=10^{3}$ Monte Carlo repeats),
the calibration is tested by $|\zeta|\le1$ and a probability of validity
is estimated as
\begin{equation}
p_{val}=\frac{1}{N}\sum_{i=1}^{N}\boldsymbol{1}(|\zeta|_{i}\le1)
\end{equation}
where $\boldsymbol{1}(x)$ is the indicator function with values 0
when $x$ is false, and 1 when $x$ is true. 

\begin{figure}[t]
\noindent \begin{centering}
\includegraphics[width=0.48\textwidth]{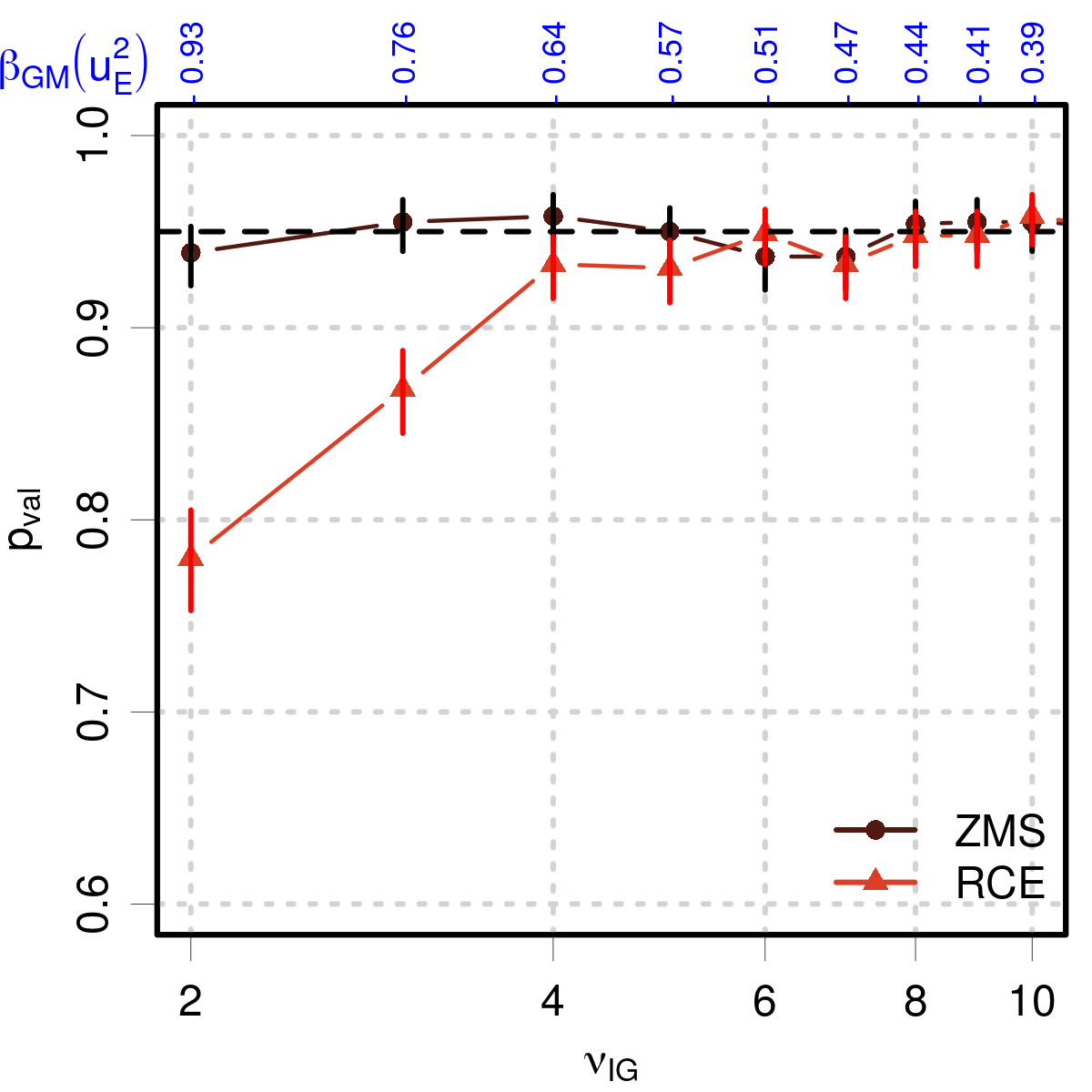}\includegraphics[width=0.48\textwidth]{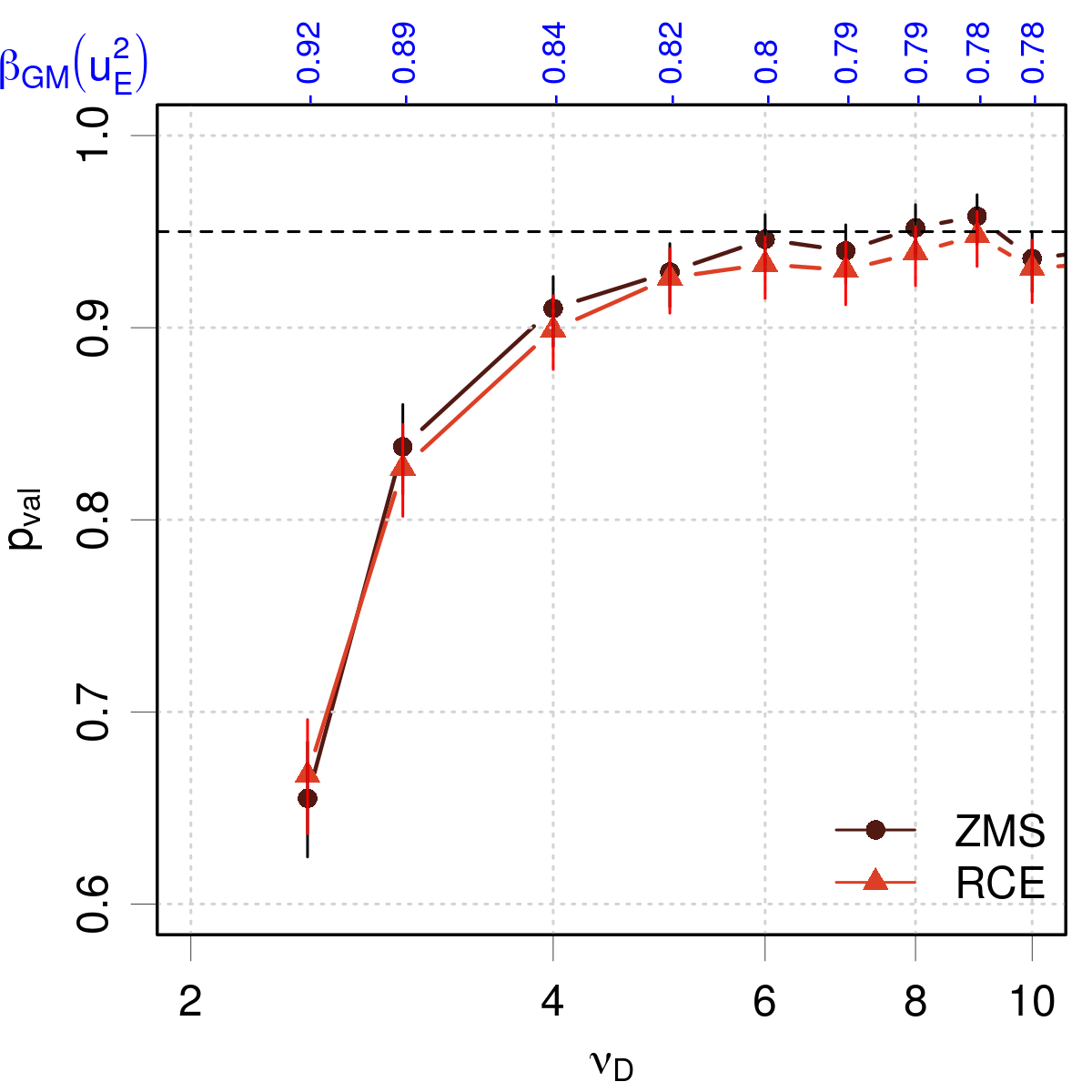}
\par\end{centering}
\caption{\label{fig:validRCE}Validation probability of the ZMS and RCE statistics
for calibrated datasets generated by two scenarios: (left) NIG with
$\nu_{IG}$ as parameter of the IG distribution; (right) TIG with
$\nu_{D}$ as parameter of the generative $D=t_{s}$ distribution.
The corresponding average values of $\beta_{GM}$ are reported on
the upper axis.}
\end{figure}

The values of $p_{val}$ and their 95\% confidence intervals obtained
by a binomial model\citep{Pernot2022a}, are plotted in Fig.\,\ref{fig:validRCE}
for the RCE and ZMS statistics as a function of $\nu_{IG}$ or $\nu_{D}$.
The upper axis provides the average $\beta_{GM}$ statistics for the
generated $u_{E}^{2}$ and $E^{2}$ samples. 

The study of the NIG model shows that the ZMS is not sensitive to
$\nu_{IG}$, even for extreme uncertainty distributions, and that
it provides intervals that consistently validate 95\,\% of the calibrated
synthetic datasets. For the RCE, the validation probability is strongly
sensitive to $\nu_{IG}$ for values below 4, reaching less than 80\,\%
for $\nu_{IG}=2$. 

When considering the TIG model, one sees that both statistics are
similarly sensitive to the shape of the generative distribution, and
that the validation intervals begin to be unreliable for $\nu_{D}<6$
($\beta_{GM}\ge0.8$). The validation probability falls to 0.65 for
$\nu_{D}=2.1$. In this case, when $\nu_{D}$ diminishes, the dispersion
of the generated $<E^{2}>$ or $<Z^{2}>$ values increases, which
is not in itself a problem, but the bootstrapped CIs are unable to
capture correctly this effect by being too narrow, i.e. too frequently
excluding the true mean value of the distribution.

For further reference one defines safety thresholds for $\beta_{GM}$.
Based on the data in Figs.\,\ref{fig:altRCE-1}-\ref{fig:validRCE},
the limit values above which one can suspect the reliability of the
calibration statistics are given in Table\,\ref{tab:limits}. Note
that these values are indicative and result from the distribution
shapes observed for the studied datasets. The values for other distribution
shapes might differ slightly. 
\begin{table}[t]
\noindent \centering{}%
\begin{tabular}{ccc}
\hline 
Variable & $\beta_{GM}$ & Impacted statistics\tabularnewline
\hline 
$u_{E}^{2}$ & 0.6 & RCE\tabularnewline
$E^{2}$, $Z^{2}$ & 0.8 & RCE, ZMS\tabularnewline
\hline 
\end{tabular}\caption{\label{tab:limits} Limits for $\beta_{GM}$ above which the reliability
of the RCE or ZMS calibration statistics should be questioned.}
\end{table}

\subsection{Summary}

This section was designed to illustrate how the calibration statistics
and their validation diagnostic are sensitive to the shape of the
uncertainty and error distributions. Two problems were identified.

The first problem is the sensitivity of the mean square (MS) statistic
$<X^{2}>$, which is central to the studied RCE and ZMS, to the presence
of outliers or heavy tails in the distribution of $X$. This is a
well known issue, which justifies often the replacement of the root
mean squared error RMSE by the more robust mean unsigned error (MUE)
in performance analysis\citep{Pernot2018,Pernot2021}. Unfortunately
the use of the MS in average calibration statistics derives directly
from the probabilistic model linking errors to uncertainties (Eq.\,\ref{eq:probmod}),
and one has to deal with its limitations, unless one is ready to change
of uncertainty paradigm, using for instance prediction intervals or
distributions.

The second problem is the inability of bootstrapping to provide reliable
confidence intervals for $<X^{2}>$ for heavy-tailed distributions. 

The robust skewness ($\beta_{GM}$) and kurtosis ($\kappa_{CS}$)
estimators were proposed and compared as tailedness metrics. Because
of their strong correlation, only $\beta_{GM}$ was retained and safety
limits have been derived to help in screening out datasets with potentially
unreliable RCE/ZMS values. It has been shown that heavy-tailed uncertainty
distributions affect mostly the RCE, while both RCE and ZMS are affected
by heavy-tailed error distributions. The ZMS benefits also from the
fact that the distribution of $Z^{2}$ has often lighter tails than
the distribution of $E^{2}$. The next section illustrates these features
on the nine example ML-UQ datasets of Sect.\,\ref{sec:ML-UQ-data}.

\section{Application to real ML-UQ datasets\label{sec:Application}}

The validation approach presented above is applied to the datasets
presented in Sec.\,\ref{sec:ML-UQ-data}. First, the datasets are
characterized by their skewness to validate the analysis of shape
parameters reported above. Then, the comparison of calibration diagnostics
for the RCE and ZMS is analyzed according to the skewness values. 

\subsection{Skewness analysis}

The first step is to assess the tailedness of the uncertainty, error
and \emph{z}-score distributions to reveal potentially problematic
datasets. The corresponding $\beta_{GM}$ values are reported in Table\,\ref{tab:data}.
\begin{table}[t]
\noindent \begin{centering}
\begin{tabular}{cr@{\extracolsep{0pt}.}lccccc}
\hline 
{\small{}Set \# } & \multicolumn{2}{c}{} & {\small{}$\beta_{GM}(u_{E}^{2})$} &  & {\small{}$\beta_{GM}(E^{2})$} &  & {\small{}$\beta_{GM}(Z^{2})$}\tabularnewline
\cline{1-1} \cline{4-4} \cline{6-6} \cline{8-8} 
{\small{}1 } & \multicolumn{2}{c}{} & {\small{}0.40} &  & \textbf{\small{}0.82}{\small{} } &  & {\small{}0.73}\tabularnewline
{\small{}2} & \multicolumn{2}{c}{} & \textbf{\small{}0.72} &  & \textbf{\small{}0.94}{\small{} } &  & \textbf{\small{}0.83}\tabularnewline
{\small{}3} & \multicolumn{2}{c}{} & \textbf{\small{}0.66} &  & {\small{}0.74 } &  & {\small{}0.69}\tabularnewline
{\small{}4} & \multicolumn{2}{c}{} & \textbf{\small{}0.74} &  & \textbf{\small{}0.82}{\small{} } &  & {\small{}0.69}\tabularnewline
{\small{}5} & \multicolumn{2}{c}{} & {\small{}0.19} &  & {\small{}0.78 } &  & {\small{}0.79}\tabularnewline
{\small{}6} & \multicolumn{2}{c}{} & {\small{}0.50} &  & \textbf{\small{}0.96}{\small{} } &  & \textbf{\small{}0.95}\tabularnewline
{\small{}7} & \multicolumn{2}{c}{} & \textbf{\small{}0.93} &  & \textbf{\small{}0.98}{\small{} } &  & {\small{}0.78}\tabularnewline
{\small{}8 } & \multicolumn{2}{c}{} & {\small{}0.30} &  & {\small{}0.79 } &  & {\small{}0.78}\tabularnewline
{\small{}9 } & \multicolumn{2}{c}{} & {\small{}0.30} &  & {\small{}0.77 } &  & {\small{}0.75}\tabularnewline
\hline 
\end{tabular}
\par\end{centering}
\caption{\label{tab:data}Robust skewness values for $u_{E}^{2}$, $E^{2}$
and $Z^{2}$. Values above the proposed safety limits (Table\,\ref{tab:limits})
are in bold type.}
\end{table}

The skewness values indicate that some distributions present heavy
tails. For uncertainties, $\beta_{GM}(u_{E}^{2})$ exceed 0.6 for
Sets 2, 3, 4, and 7. This list conforms to the shape analysis of Table\,\ref{tab:dataNu},
except for Set 6, which had a very small $\nu$ parameter but has
a moderate skewness. For the errors, the largest values occur for
Sets 1, 2, 4, 6 and 7, a list where the shape analysis of Table\,\ref{tab:dataNu}
would have added Set 5. For the \emph{z}-scores, only two sets are
screened out, Sets 2 and 6, while Sets 5 and 6 have small $\nu$ values.
The correspondence between the previous shape analysis and the skewness
screening is not perfect, which reflects the imperfect quality of
the distributions fits provided by the IG and F models. According
to the present analysis, Sets 1, 2, 3, 4, 6 and 7 are potentially
problematic, Set 1 being very close to the limit.

\begin{figure}[t]
\noindent \begin{centering}
\includegraphics[width=0.6\textwidth]{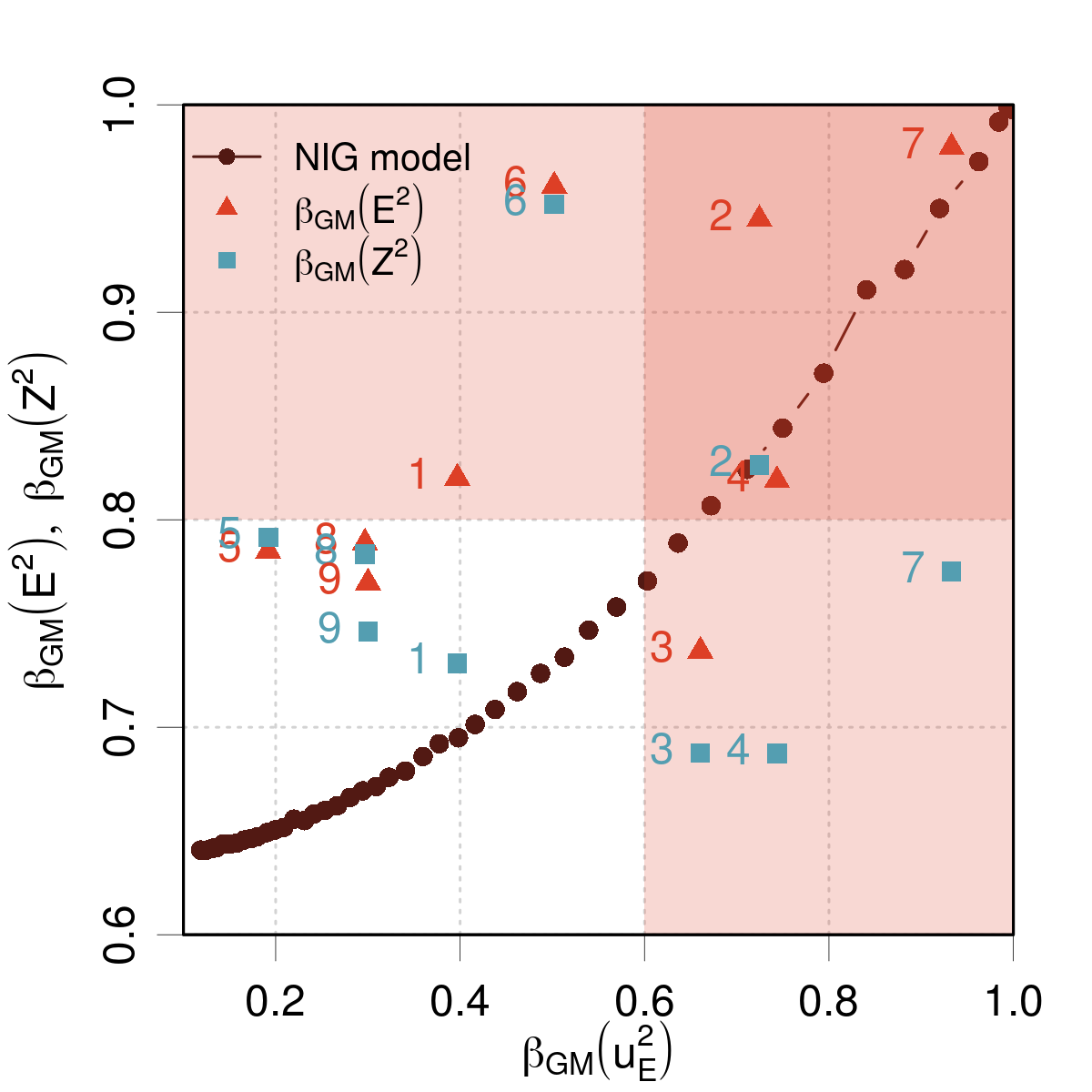}
\par\end{centering}
\caption{\label{fig:validBetaKappa}Skewness analysis of the application datasets.
The black dots figure the NIG model for $2\le\nu\le20$. The colored
areas signal the values that might be associated with problems.}
\end{figure}

A graphical summary of the skewness analysis is provided in Fig.\,\ref{fig:validBetaKappa},
where $\beta_{GM}(E^{2})$ is plotted against $\beta_{GM}(u_{E}^{2})$
for the nine datasets (red triangles) and for the NIG model ($2\le\nu\le20$;
black dots). One sees directly that only Sets 5, 8 and 9 are in the
safe area for both statistics, although they lie close to the $E^{2}$
skewness limit. All the other sets have a skewness value that might
be problematic for at least one of $u_{E}^{2}$ or $E^{2}$. It is
also noteworthy that most $\beta_{GM}(E^{2})$ values (except for
Sets 3, 4 and 7) exceed what should be expected from the NIG model,
confirming the shape analysis of Sect.\,\ref{sec:ML-UQ-data}. 

The blue squares depict the values for $\beta_{GM}(Z^{2})$: except
for Sets 2 and 6, they all lie below the safety limit (0.8), and for
Set 2 the value is much closer to the limit than $\beta_{GM}(E^{2})$.
This indicates that the ZMS is much less likely to be affected by
estimation problems than the RCE, but might still be challenged by
data such as Sets 2 and 6. 

\subsection{Comparison of validation scores\label{subsec:Comparison-of-validation}}

The statistics and bootstrapped confidence intervals have been estimated
for the RCE and ZMS for all datasets with $10^{4}$ bootstrap replicates.
The resulting values and validation scores are reported in Table\,\ref{tab:RCE-statistic-and}.
It is clear from these results that average calibration is not satisfied
by all datasets and, more problematically, that the diagnostic depends
notably on the choice of statistic. 
\begin{table}[t]
\noindent \begin{centering}
\begin{tabular}{r@{\extracolsep{0pt}.}lr@{\extracolsep{0pt}.}lr@{\extracolsep{0pt}.}lcr@{\extracolsep{0pt}.}lr@{\extracolsep{0pt}.}lr@{\extracolsep{0pt}.}lr@{\extracolsep{0pt}.}lr@{\extracolsep{0pt}.}lr@{\extracolsep{0pt}.}lr@{\extracolsep{0pt}.}l}
\hline 
\multicolumn{9}{l}{$\vartheta=RCE$; $\vartheta_{ref}=0$} & \multicolumn{2}{c}{} & \multicolumn{10}{l}{$\vartheta=ZMS$; $\vartheta_{ref}=1$}\tabularnewline
\cline{1-9} \cline{12-21} 
\multicolumn{2}{c}{Set } & \multicolumn{2}{c}{$\vartheta_{est}$} & \multicolumn{2}{c}{$b_{BS}$} & $I_{BS}$ & \multicolumn{2}{c}{$\zeta_{RCE}$} & \multicolumn{2}{c}{} & \multicolumn{2}{c}{Set } & \multicolumn{2}{c}{$\vartheta_{est}$} & \multicolumn{2}{c}{$b_{BS}$} & \multicolumn{2}{c}{$I_{BS}$} & \multicolumn{2}{c}{$\zeta_{ZMS}$}\tabularnewline
\cline{1-9} \cline{12-21} 
\multicolumn{2}{c}{1 } & 0&019 & 2&2e-04  & {[}-0.021, 0.055{]} & \textbf{0}&\textbf{47} & \multicolumn{2}{c}{} & \multicolumn{2}{c}{1 } & 0&96 & -2&7e-04  & {[}0&87, 1.11{]} & \textbf{-0}&\textbf{27}\tabularnewline
\multicolumn{2}{c}{2 } & -0&039 & 5&9e-04  & {[}-0.106, 0.020{]} & \textbf{-0}&\textbf{66} & \multicolumn{2}{c}{} & \multicolumn{2}{c}{2 } & 0&89 & -7&3e-04  & {[}0&80, 0.999{]} & -1&01\tabularnewline
\multicolumn{2}{c}{3 } & -0&0075 & -2&1e-04  & {[}-0.054, 0.040{]} & \textbf{-0}&\textbf{16} & \multicolumn{2}{c}{} & \multicolumn{2}{c}{3 } & 1&12 & 9&4e-05  & {[}1&05, 1.2{]} & 1&73\tabularnewline
\multicolumn{2}{c}{4 } & 0&055 & -3&9e-04  & {[}-0.0025, 0.12{]} & \textbf{0}&\textbf{96} & \multicolumn{2}{c}{} & \multicolumn{2}{c}{4 } & 1&23 & -6&4e-04  & {[}1&16, 1.3{]} & 3&50\tabularnewline
\multicolumn{2}{c}{5 } & 0&099 & 1&4e-04  & {[}0.057, 0.14{]} & 2&33 & \multicolumn{2}{c}{} & \multicolumn{2}{c}{5 } & 0&85 & 1&3e-04  & {[}0&78, 0.93{]} & -1&84\tabularnewline
\multicolumn{2}{c}{6 } & 0&092 & 9&9e-04  & {[}0.00079, 0.16{]} & 1&01 & \multicolumn{2}{c}{} & \multicolumn{2}{c}{6 } & 0&98 & -6&2e-04  & {[}0&85, 1.15{]} & \textbf{-0}&\textbf{10}\tabularnewline
\multicolumn{2}{c}{7 } & -0&26 & 5&5e-03  & {[}-0.68, -0.0012{]} & -1&00 & \multicolumn{2}{c}{} & \multicolumn{2}{c}{7 } & 0&97 & 2&6e-04  & {[}0&94, 1.01{]} & \textbf{-0}&\textbf{69}\tabularnewline
\multicolumn{2}{c}{8 } & 0&046 & 1&3e-05  & {[}0.0082, 0.077{]} & 1&22 & \multicolumn{2}{c}{} & \multicolumn{2}{c}{8 } & 0&93 & 3&4e-04  & {[}0&87, 0.99{]} & -1&12\tabularnewline
\multicolumn{2}{c}{9 } & -0&013 & 2&1e-04  & {[}-0.072, 0.027{]} & \textbf{-0}&\textbf{33} & \multicolumn{2}{c}{} & \multicolumn{2}{c}{9 } & 0&97 & 2&5e-04  & {[}0&90, 1.08{]} & \textbf{-0}&\textbf{26}\tabularnewline
\hline 
\end{tabular}
\par\end{centering}
\caption{\label{tab:RCE-statistic-and}RCE and ZMS statistics and their validation
results. The bold $\zeta_{x}$ values indicate calibrated sets, where
$|\zeta_{x}|\le1$.}
\end{table}

Comparison of the absolute $\zeta$-scores for ZMS and RCE across
the nine datasets shows a contrasted situation {[}Fig.\,\ref{fig:discrepancy}(left){]}:
\begin{figure}[t]
\noindent \begin{centering}
\includegraphics[width=0.48\textwidth]{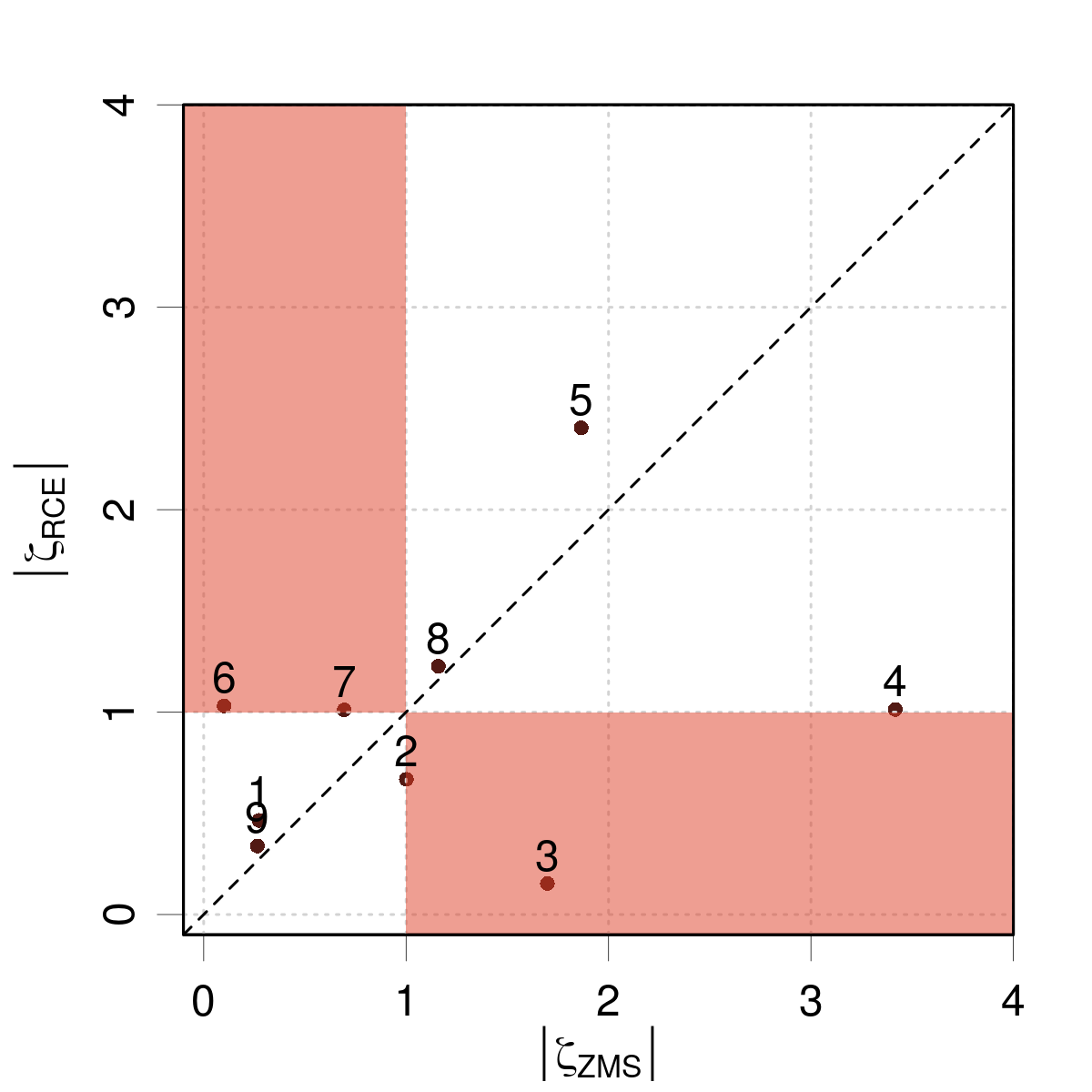}\includegraphics[width=0.48\textwidth]{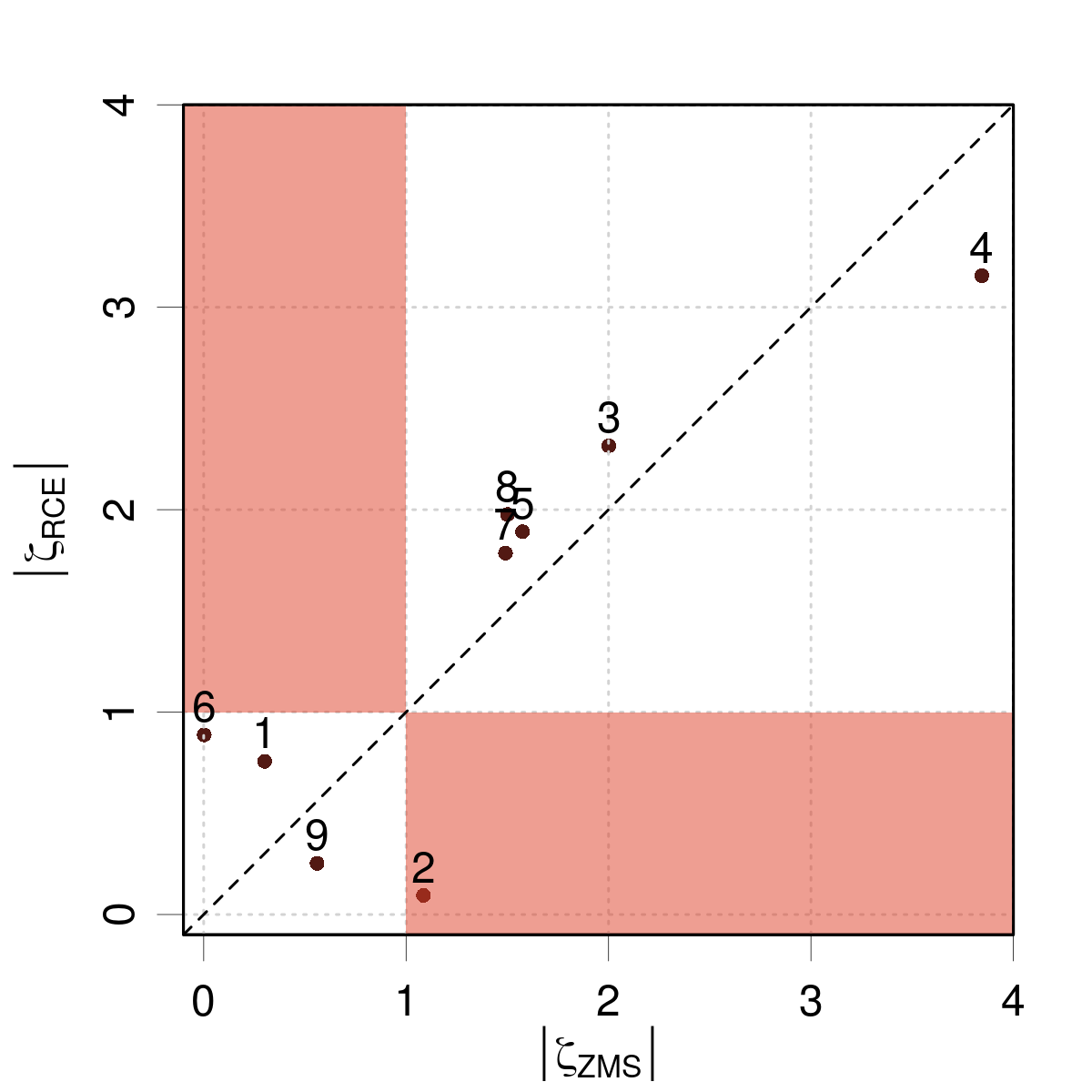}
\par\end{centering}
\caption{\label{fig:discrepancy}Comparison of the absolute $\zeta$-scores
for ZMS and RCE: (left) the original datasets; (right) after removal
of the 5\,\% largest uncertainties. The symbols represent the set
numbers in Table\,\ref{tab:RCE-statistic-and}. The colored areas
contain the disagreeing validation results.}
\end{figure}

\begin{itemize}
\item Points close to the identity line and more globally in uncolored areas,
are the datasets for which both statistics agree on the calibration
diagnostic, i.e. positive for Sets 1 and 9, and negative for Set 5
and 8. 
\item Sets 6 and 7 are validated by the ZMS and rejected by the RCE, although
very close to the limit.
\item Finally, Sets 2, 3 and 4 are validated by RCE and rejected by ZMS. 
\end{itemize}
Globally, the statistics disagree on more than half of the datasets,
which is surprising for two statistics deriving analytically from
the same generative model (Eq.\,\ref{eq:probmod}). Considering that
the RCE ignores the pairing of uncertainties and errors, one could
have expected it to be more forbidding than the ZMS, which is the
case for Sets 2, 3 and 4, but not for Set 6 (maybe also for Sets 5
and 7). 

It is remarkable that the five sets for which a disagreement between
RCE and ZMS is observed (Sets 2, 3, 4, 6 and 7) are those with the
largest skewness values for $u_{E}^{2}$ (Table\,\ref{tab:data},
Fig.\,\ref{fig:validBetaKappa}), with $\beta_{GM}(u_{E}^{2})\ge0.6$,
except for Set 6, as shown in Fig.\,\ref{fig:validBetaKappa}(left).
This suggests that the upper tail of the uncertainty distribution
plays a major role in this disagreement. Sets 2, 4, 6 and 7 present
also $\beta_{GM}(E^{2})\ge0.8$, which might point to problematic
error distributions.

The impact of the tails of the uncertainty and error distributions
is assessed in the next section.

\subsection{Impact of the distributions tails\label{sec:Analysis}}

In order to better understand the discrepancy of the validation results
by RCE and ZMS, one needs to consider the sensitivity of these statistics
to the uncertainty distributions, and notably to the large, sometimes
outlying, values, as suggested by the skewness analysis. In the \emph{z}-scores,
these large uncertainty values are likely to contribute to small absolute
values of $Z$ having a small impact on the ZMS, while they are likely
to affect significantly the estimation of the RCE. This hypothesis
is tested on the nine datasets by a decimation experiment, where both
statistics are estimated on datasets iteratively pruned from their
largest uncertainties.

\begin{figure}[t]
\noindent \begin{centering}
\includegraphics[width=0.95\textwidth]{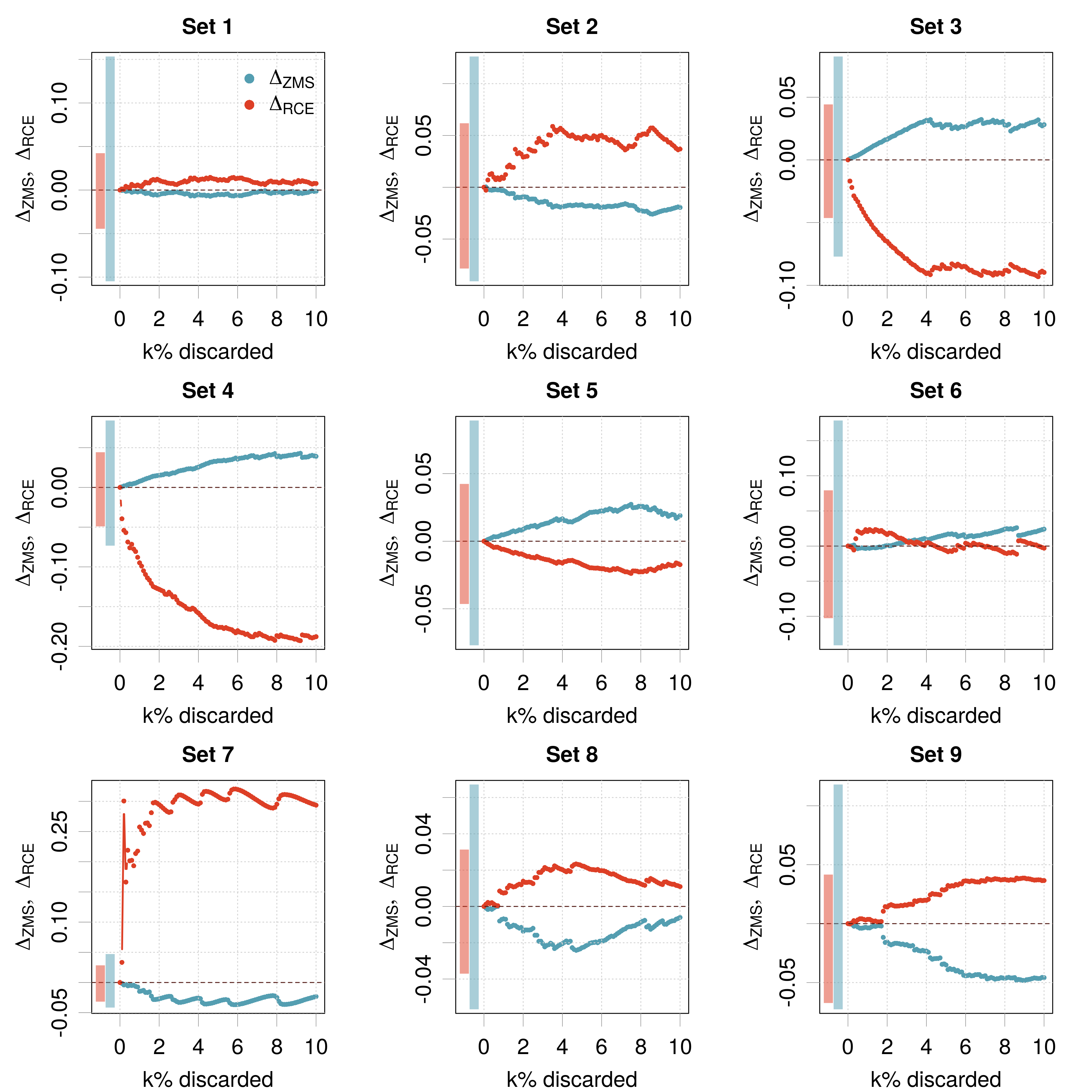}
\par\end{centering}
\caption{\label{fig:decimation}Variation of the RCE and ZMS statistics according
to the percentage $k$ of largest uncertainties removed from the datasets.
The vertical colored bars represent 95\,\% confidence intervals on
the statistics for the full datasets.}
\end{figure}

The deviations of the ZMS and RCE scores from their value for the
full dataset are estimated for an iterative pruning (decimation) of
the datasets from their largest uncertainties, as performed in confidence
curves\citep{Pernot2022c}. The values of $\Delta_{RCE}=RCE_{k}-RCE_{0}$
and $\Delta_{ZMS}=ZMS_{k}-ZMS_{0}$ for a percentage $k$ of discarded
data varying between 0 and 10\,\% are shown in Fig.\,\ref{fig:decimation},
where zero-centered bootstrapped 95\,\% CIs for both statistics are
displayed as vertical bars to assess the amplitude of the deviations.

It appears that in all the cases ZMS is less or as sensitive as RCE
to the large extremal uncertainty values and that its decimation curve
always strays within the limits of the corresponding CI. For RCE,
one can find cases where it is more sensitive than ZMS but lies within
the limits of the CI (Sets 2 and 6) and cases where it strays beyond
the limits of the CI (Sets 3, 4 and 7). These five cases are precisely
those where the RCE diagnostic differs the most from the ZMS (Sect.\,\ref{subsec:Comparison-of-validation}).
This sensitivity test confirms the hypothesis that RCE is more sensitive
than ZMS to the upper tail of the uncertainty distribution, to a point
where its estimation might become unreliable.

For some sets (2, 3, 6 and 7) a small percentage (1-4\,\%) of the
largest uncertainties affects the $\Delta_{RCE}$ statistic, after
which one observes a sharp change of slope, and the curves becomes
a reflected image of the $\Delta_{ZMS}$ statistic. These uncertainties
might therefore be considered as outlying values. By contrast, one
observes for Set 4 a smooth transition, which suggest that the shape
of the tail is involved, rather than a few extreme uncertainty values.

The sensitivity of the $\zeta$-scores to the removal of the 5\,\%
largest uncertainties is shown in Fig.\,\ref{fig:discrepancy}(right).
One can expect from this truncation a better agreement of ZMS and
RCE for those datasets with heavy $u_{E}$ tails. The agreement is
indeed notably improved for Set 3, 4 and 7. 

Surprisingly, the $\zeta_{RCE}$ values for Sets 2 and 6 do not follow
the same trend, with a deterioration for Set 2 and no notable change
for Set 6. Note that these are the sets for which the $\Delta_{RCE}$
curve does \emph{not} stray notably out of the corresponding CI limits.
The main reason for the discrepancy of validation metrics for Sets
2 and 6 is therefore not the shape of the uncertainty distributions,
which leads us to the error distributions. 

Both datasets present very large error skewness values $\beta_{GM}(E^{2})$
(0.92 and 0.96) , well above the safety limit of 0.8. The skewness
of z-scores $\beta_{GM}(Z^{2})$ for these sets (0.83 and 0.95), although
smaller than $\beta_{GM}(E^{2})$, are also above the safety limit.
It is therefore difficult to conclude on the average calibration of
Sets 2 and 6, as both the RCE and ZMS are likely to be affected by
the heavy-tails of the $E^{2}$ and $Z^{2}$ distributions, respectively. 

\subsection{Additional datasets\label{subsec:Additional-datasets}}

After the completion of this study, Jacobs \emph{et al.}\citep{Jacobs2024}
published an ensemble of 33 datasets of ML materials properties, with
predictions by random forest models and calibrated uncertainty estimates.
These new datasets offer a unique opportunity to test the generalizability
of the analysis presented above. For this, one will first analyze
the skewness of the $Z^{2}$ distributions and then observe the discrepancy
of the RCE and ZMS calibration statistics. 

The prediction uncertainties in these datasets have been calibrated
\emph{pos-hoc,} by a polynomial transformation. The parameters of
the calibration polynomial have been estimated by NLL minimization,
which, in this setup, is equivalent to an optimization of the ZMS
to its target value (Eq.\,\ref{eq:NLL}). It is therefore expected
that all the datasets have good ZMS statistics. Skewness and calibration
statistics are reported in Table \ref{tab:Stats-1}-\ref{tab:Stats2-1}
of Appendix\,\ref{sec:Statistics-of-Jacobs}. Due to aberrant data
and statistics, Set 13 is reported, but excluded from the discussions.
\begin{figure}[t]
\noindent \begin{centering}
\includegraphics[width=0.6\textwidth]{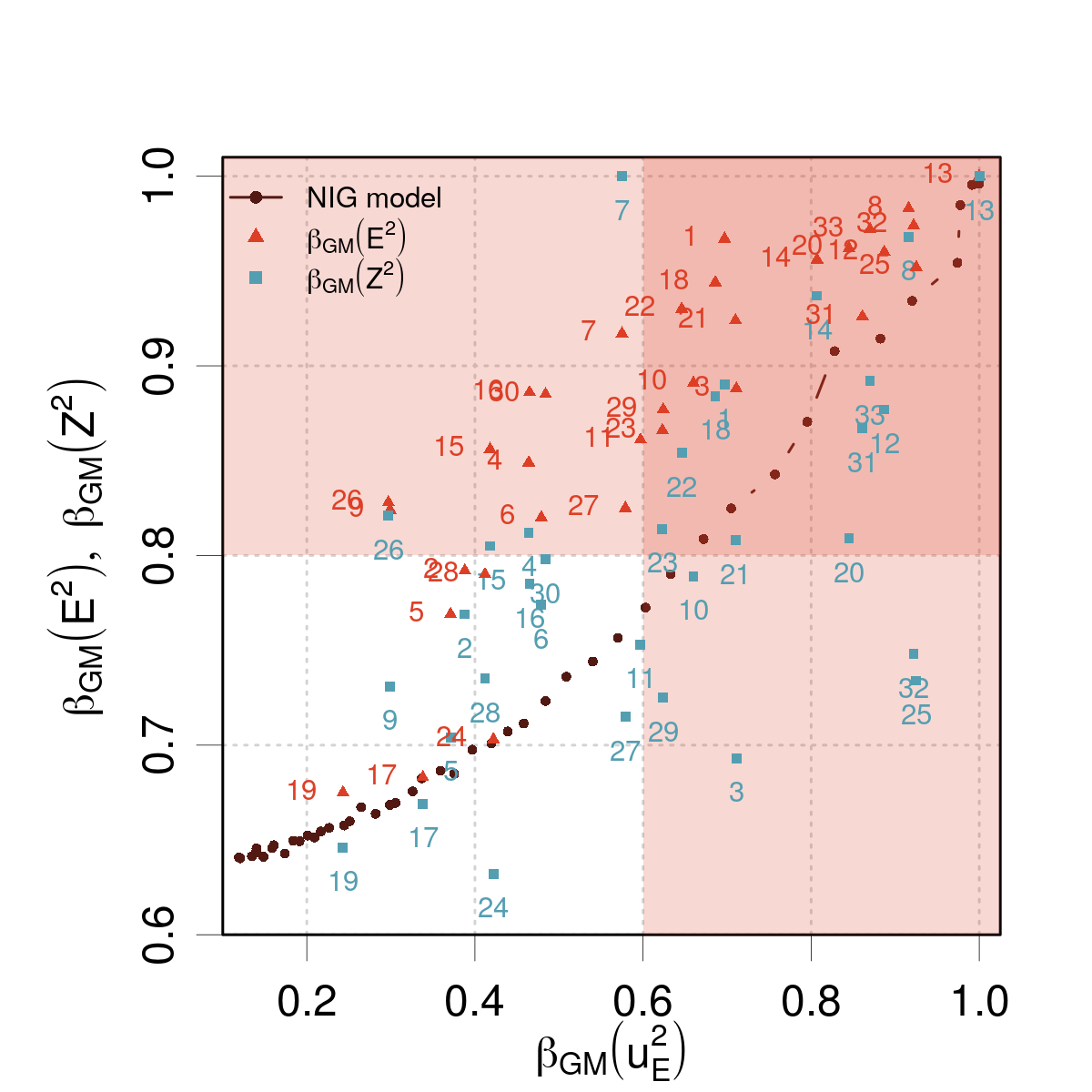}
\par\end{centering}
\caption{\label{fig:skewness}Same as Fig.\,\ref{fig:validBetaKappa} for
Jacobs \emph{et al.}\citep{Jacobs2024} 33 datasets.}
\end{figure}

\subsubsection{Skewness analysis}

All datasets in the colored areas in Fig.\,\ref{fig:skewness} are
likely to have calibration statistics influenced by heavy tails and/or
outliers: about half of the datasets have heavy-tailed $u_{E}^{2}$
distributions, while more than 5 out of 6 have heavy $E^{2}$ tails
(red triangles); many of them present both features. Only 6 datasets
are in the ``safe'' area where the estimation of mean squared statistics
should not be too biased (2, 5, 17, 19, 24, 28). It is also remarkable
that a few sets (17, 19, 24, 25 and 31) lie close to the NIG model
line, indicating that they might be associated with a quasi-normal
generative distribution. However, most of the points lie far above
the NIG model line, showing that their generative distributions are
far from being normal. Here again, the $Z^{2}$ distributions (blue
squares) are globally less problematic than the error distributions,
but only a little more than half of the cases lies below the $\beta_{GM}(Z^{2})=0.8$
threshold. Set 7 presents an awkward scenario where the skewness of
$Z^{2}$ is larger than the skewness of $E^{2}$, certainly due to
the presence of outlying error values, as noted in Appendix\,\ref{sec:Statistics-of-Jacobs}.
Only two sets (Sets 19 and 24) have $\beta_{GM}(Z^{2})$ values compatible
with a normal distribution (0.64). This larger panel of cases confirms
the previous observations that many ML-UQ datasets present distributions
that might be challenging for the reliability of variance-based calibration
statistics. 

In their analysis, Jacobs \emph{et al.}\citep{Jacobs2024} associate
calibration with the standard-normality of \emph{z}-scores (see for
instance Sects.\,2.3 and 3.2 of their article). If this were the
case, very few of the 33 datasets should be considered as calibrated.
In fact, and according to the generative model (Eq.\,\ref{eq:probmod}),
calibration is assessed by unbiasedness ($<Z>=0$) and unit ZMS ($<Z^{2}>=1$)
or unit variance ($\mathrm{Var}(Z$)=1), without reference to the
shape of the distribution. The NLL score minimized for the calibration
of uncertainties cannot ensure the normality of \emph{z}-scores, because
it does not involve nor constrain moments higher than the second order.

\begin{figure}[t]
\noindent \begin{centering}
\includegraphics[width=0.48\textwidth]{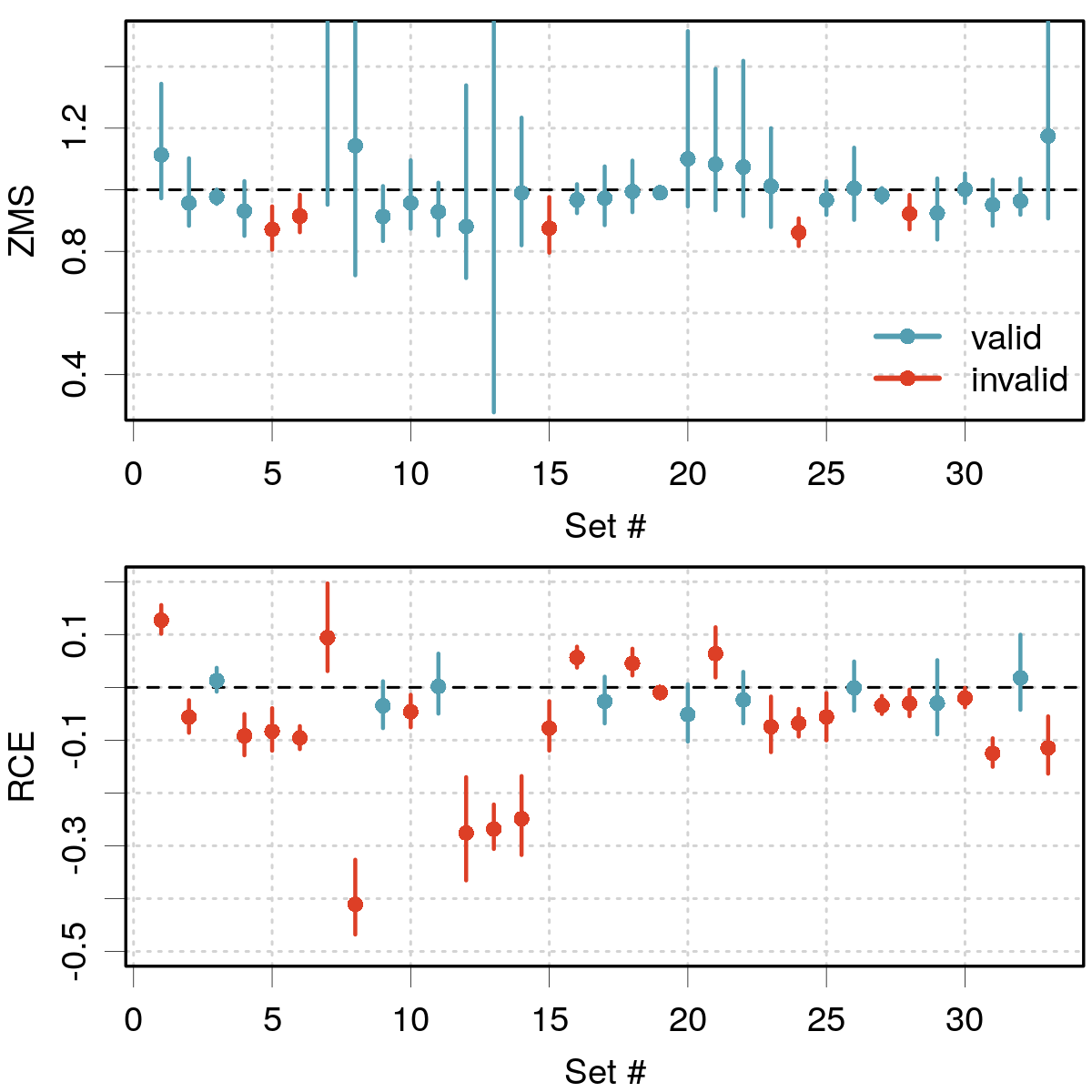}\includegraphics[width=0.48\textwidth]{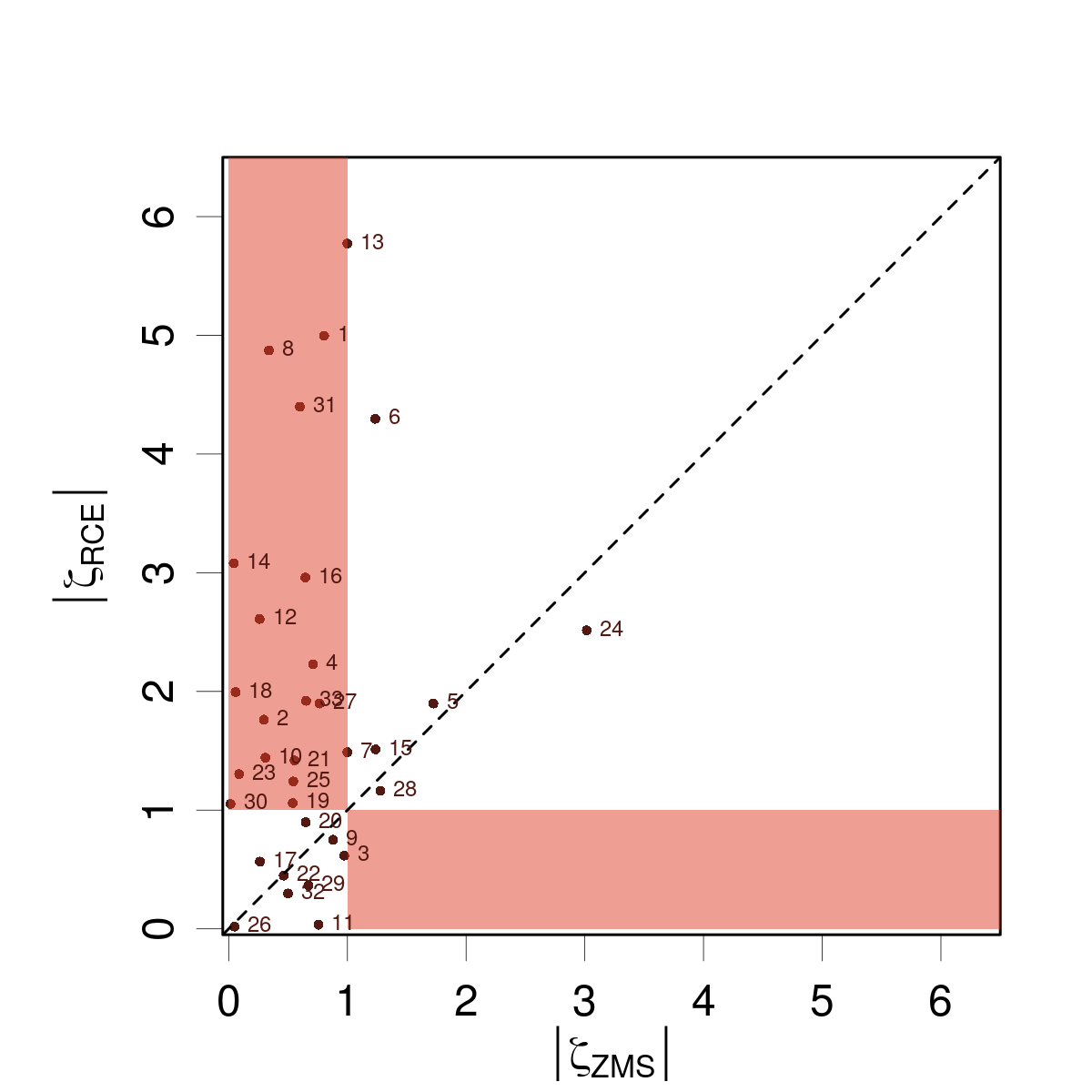}
\par\end{centering}
\caption{\label{fig:valid-1}Comparison of ZMS and RCE calibration and validation
scores for Jacobs \emph{et al.}\citep{Jacobs2024} 33 datasets. (left)
ZMS and RCE calibration statistics with bootstrapped 95\,\% confidence
intervals. Red intervals do not cover the target value. (right) Same
as Fig.\,\ref{fig:discrepancy} .}
\end{figure}

\subsubsection{Comparison of calibration and validation scores}

Fig.\,\ref{fig:valid-1}(left) presents the ZMS and RCE calibration
statistics with their bootstrapped 95\% confidence intervals. Considering
the ZMS, one sees that Sets 7 (and 13), despite huge ZMS values, are
validated due to very wide confidence intervals. This raises an alert
on the blind use of validation statistics such as $\zeta$-scores
for heavy-tailed distributions or datasets with outliers. One can
also see that Sets 5, 6, 15, 24 and 28 have invalidated ZMS values.
These datasets should therefore not be considered as calibrated. 

The RCE values show a much larger panel of invalidated datasets. This
discrepancy between ZMS and RCE is problematic and reflects the unreliability
of these statistics for heavy-tailed uncertainty and/or error distributions.
The statistics agree for Sets 3, 5, 6, 9, 11, 15, 17, 20, 22, 24,
26, 28, 29 and 32, i.e. less than half of the cases (14 over 33).
The majority of these agreements correspond to datasets with $\beta_{GM}(Z^{2})\le0.8$.
It is therefore legitimate to question the validity of the calibration
statistics for the other, heavy-tailed, datasets. 

Another way to look at the disagreement of the RCE and ZMS statistics
is to compare their zeta scores. Fig.\,\ref{fig:valid-1}(right)
comforts the skewness analysis. Points close to the diagonal are those
for which the calibration diagnostics (either positive or negative)
agree. Most of the sets in the safe area for skewness (Fig.\,\ref{fig:skewness})
fall in this case (5, 17, 24 and 28). But, globally, the calibration
statistics disagree, as observed also in Fig.\,\ref{fig:valid-1}(left).

\section{Conclusions\label{sec:Conclusion}}

This study used the RCE and ZMS average calibration statistics to
illustrate the impact of heavy-tailed uncertainty and/or error distributions
on the reliability of calibration statistics. These statistics have
been used because they have well defined reference values for statistical
validation, which is not the case of the more popular bin-based statistics,
such as the ENCE.

Discrepancies of validation diagnostic was observed between both statistics
for an ensemble of datasets extracted from the recent ML-UQ literature
for regression tasks in materials and chemical sciences. This anomaly
has been elucidated by showing that the RCE is very sensitive to the
upper tail of the uncertainty distribution, and notably to the presence
of outliers, which has been proven by a decimation experiment (5 cases
out of 9). Moreover, error sets with heavy tails can also be challenging
for the RCE (5 cases). The distribution of z-scores ($Z=E/u_{E}$)
appeared to be less problematic, although some heavy-tailed $Z$ distributions
have been observed (2 cases). 

Two underlying problems contribute to this state of affairs: (1) strong
sensitivity of the Mean Squares statistic to the presence of heavy-tails
and outliers; and (2) our inability to design reliable confidence
intervals with a prescribed coverage for the mean of variables with
heavy-tailed distributions ($u_{E}^{2}$, $E^{2}$ and $Z^{2}$).
Bootstrapping is presently the best approach to CI estimation for
the mean of non-normal variables, and numerical experiments have shown
that it fails in conditions representative of the studied datasets. 

A major consequence of these observations is that average calibration
statistics based on the comparison of $MSE=<E^{2}>$ to $MV=<u_{E}^{2}>$
should not be blindly relied upon for the kind of datasets found in
ML-UQ regression problems. The RCE is affected by both MSE and MV
sensitivity to heavy tails. In contrast, the ZMS statistic has fewer
reliability issues and should therefore be the statistic of choice
for average calibration testing. Nevertheless, it should not be expected
to be fully reliable.

In this context, it is important to be able to screen out problematic
datasets. It has been shown that a robust skewness statistic ($\beta_{GM}$)
can be used for this, and safety thresholds have been defined for
the $u_{E}^{2}$, $E^{2}$ and $Z^{2}$ distributions. The reliability
of the RCE and ZMS statistics for datasets with values exceeding these
limits has to be questioned.

A temptation would be to ignore altogether average calibration statistics
and to focus on conditional calibration. As the RCE, the popular UCE
and ENCE statistics implement the comparison of \emph{MV} to \emph{MSE},
but, being bin-based, one might expect them to be less susceptible
to the above-mentioned sensitivity issue when the binning variable
is $u_{E}$ (\emph{consistency} testing\citep{Pernot2023d}). Unfortunately,
even if binning solves the tailedness problem for $<u_{E}^{2}>$ for
most of the bins, this is not the case for $<E^{2}>$, leaving these
conditional calibration statistics with the same reliability problem
as the RCE. This also applies to the local ZMS analysis. The problem
of conditional calibration statistics will be further explored in
a forthcoming article.

In \emph{post hoc} calibration, scaling factors for the uncertainties
can be derived from the ZMS ($\sigma$ scaling\citep{Laves2020},
BVS\citep{Frenkel2023,Pernot2023c_arXiv}, polynomial scaling\citep{Palmer2022,Jacobs2024})
or from the UCE\citep{Frenkel2023}. So, by extension, the outlined
problems also affect \emph{post-hoc} calibration procedures based
on statistics related to the RCE, ENCE or ZMS (for instance the NLL).
Those methods parameterized on datasets with heavy uncertainty and/or
error tails might also be unreliable.

An ensemble of 33 datasets\citep{Jacobs2024}, published after the
completion of this study, has been included to validate these conclusions
and it fully confirms them. One can therefore be confident that heavy
tailed distribution of uncertainties, errors and z-scores are presently
a notable feature of ML materials properties datasets which has to
be seriously taken into account.

\medskip{}

One can envision three paths out of this lack of reliability of calibration
statistics: 
\begin{enumerate}
\item The reduction of the problematic tails of uncertainty and error distributions
can be handled to some extent by iterative learning, i.e. by iteratively
feeding the data with the largest absolute errors and uncertainties
to the training algorithm until the problem subsides\citep{Pernot2020b}.
\item For datasets with uncontrollable heavy tails, one might have to abandon
formal validation and to rely on graphical methods to derive a qualitative
calibration diagnostic\citep{Pernot2022a,Pernot2022b}.
\item A change of uncertainty metric, using prediction intervals or distributions
instead of standard deviations, would enable a better handling of
heavy tails and seems presently to be the most promising alternative
to recover a sound statistical validation framework for calibration.
\end{enumerate}

\section*{Acknowledgments}

\noindent I warmly thank J. Busk for providing the QM9 dataset and
R. Jacobs for his help with the ``33 datasets''.

\section*{Author Declarations}

\subsection*{Conflict of Interest}

The author has no conflicts to disclose.

\section*{Code and data availability\label{sec:Code-and-data}}

\noindent The code and data to reproduce the results of this article
are available at \url{https://github.com/ppernot/2024_RCE/releases/tag/v4.0}
and at Zenodo (\url{https://doi.org/10.5281/zenodo.13341150}). The
33 datasets of Jacobs \emph{et al.}\citep{Jacobs2024} are accessible
in a FigShare depository\citep{Morgan2024_FigShare}. 

\bibliographystyle{unsrturlPP}
\bibliography{NN}

\clearpage{}

\appendix

\section{An alternative formulation of RCE\label{sec:An-alternative-formulation}}

The biased estimation of the RCE as defined by Eq.\,\ref{par:RCE}
for heavy-tailed uncertainty and error distributions can be reduced
by expressing it as
\begin{equation}
RCE2=\frac{MV-MSE}{MV}
\end{equation}
where the mean variance (MV) and mean squared errors (MSE) are used
instead of their square roots.

\begin{figure}[t]
\noindent \begin{centering}
\includegraphics[width=0.48\textwidth]{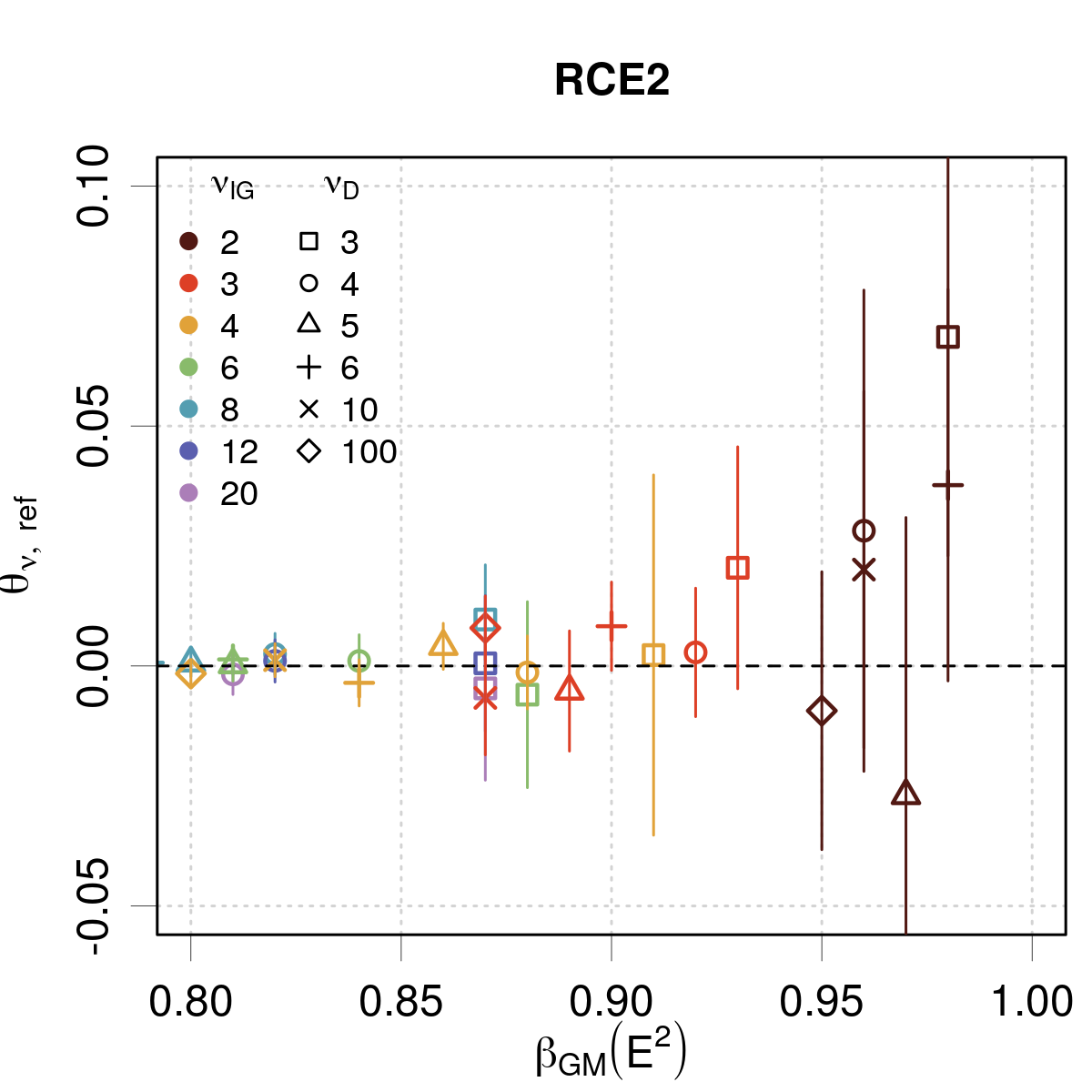} 
\par\end{centering}
\caption{\label{fig:altRCE}Estimated values of an alternative RCE formulation
for a series of datasets generated by TIG models and characterized
by the error skewness parameter $\beta_{GM}(E^{2})$.}
\end{figure}

Simulations with the same setup as in Sect.\,\ref{subsec:Sensitivity-of-calibration}
were performed for the RCE2 statistic. The results are reported in
Fig.\,\ref{fig:altRCE}.

When compared to the results for the original RCE {[}Fig.\,\ref{fig:altRCE-1}(left){]},
two features are remarkable: (1) the mean values present also deviations
increasing with $\beta_{GM}$, but with positive or negative signs;
and (2) the Monte Carlo $2\sigma$ error bars now cover the reference
value, so that there is no systematic bias for RCE2.

\clearpage{}

\section{Statistics of Jacobs \emph{et al.} datasets\label{sec:Statistics-of-Jacobs}}

Skewness and calibration statistics for the 33 datasets of \emph{Jacobs
et al.} \citep{Jacobs2024} are presented in Table \ref{tab:Stats-1}-\ref{tab:Stats2-1}. 

The uncertainties $u_{E}$ are taken from the \texttt{model\_error\_leaveout\_calibrated.csv}
files, and the errors $E$ from the \texttt{residuals\_leavout.csv}
files. Note that a few datasets present anomalous data, such as non-positive
uncertainties, perfectly null errors or remarkable outliers:
\begin{itemize}
\item Set 1 has 10\% of null errors;
\item Set 6 has 1 non-positive uncertainty;
\item Set 7 has 5 outlying negative errors leading to a very large ZMS value;
\item Set 13 has many non-positive uncertainties and null errors. 
\end{itemize}
Null errors might result from the prediction of properties with physical
limits (such as band gaps that should be positive or null). Their
occurrence is only problematic for the logarithmic representation
of error or \emph{z}-scores distributions, such as in Figs\,\ref{fig:fitE2}.
These data are \emph{not} excluded from the statistical analysis.
Similarly, the outliers of Set 7 have been kept. \emph{A contrario},
non-positive uncertainties are most likely an artifact of an unconstrained
NLL calibration procedure, and they have systematically been filtered-out
for the processing of these datasets. Note that for Set 13, the ZMS
statistic is numerically aberrant, even after filtering.

\begin{table}[t]
\begin{tabular}{llr@{\extracolsep{0pt}.}lr@{\extracolsep{0pt}.}lr@{\extracolsep{0pt}.}lr@{\extracolsep{0pt}.}lr@{\extracolsep{0pt}.}lr@{\extracolsep{0pt}.}lr@{\extracolsep{0pt}.}lr@{\extracolsep{0pt}.}lr@{\extracolsep{0pt}.}lr@{\extracolsep{0pt}.}l}
\hline 
\# & Name  & \multicolumn{2}{c}{$M$} & \multicolumn{2}{c}{$n_{0}(u_{E})$ } & \multicolumn{2}{c}{$n_{0}(E)$} & \multicolumn{2}{c}{} & \multicolumn{2}{c}{$\beta_{GM}(u_{E}^{2})$} & \multicolumn{2}{c}{$\beta_{GM}(E^{2})$} & \multicolumn{2}{c}{$\beta_{GM}(Z^{2})$} & \multicolumn{2}{c}{} & \multicolumn{2}{c}{ZMS } & \multicolumn{2}{c}{RCE}\tabularnewline
\hline 
1 & bandgap\_expt  & \multicolumn{2}{c}{30150 } & \multicolumn{2}{c}{0 } & \multicolumn{2}{c}{3070 } & \multicolumn{2}{c}{} & \textbf{0}&\textbf{7 } & \textbf{0}&\textbf{97 } & \textbf{0}&\textbf{89 } & \multicolumn{2}{c}{} & 1&11  & \textbf{0}&\textbf{13}\tabularnewline
2 & concrete  & \multicolumn{2}{c}{5150 } & \multicolumn{2}{c}{0 } & \multicolumn{2}{c}{0 } & \multicolumn{2}{c}{} & 0&39  & 0&79  & 0&77  & \multicolumn{2}{c}{} & 0&96  & \textbf{-0}&\textbf{06}\tabularnewline
3 & debyeT\_aflow  & \multicolumn{2}{c}{24480 } & \multicolumn{2}{c}{0 } & \multicolumn{2}{c}{0 } & \multicolumn{2}{c}{} & \textbf{0}&\textbf{71 } & \textbf{0}&\textbf{89 } & 0&69  & \multicolumn{2}{c}{} & 0&98  & 0&01\tabularnewline
4 & dielectric  & \multicolumn{2}{c}{5280 } & \multicolumn{2}{c}{0 } & \multicolumn{2}{c}{0 } & \multicolumn{2}{c}{} & 0&46  & \textbf{0}&\textbf{85 } & \textbf{0}&\textbf{81 } & \multicolumn{2}{c}{} & 0&93  & \textbf{-0}&\textbf{09}\tabularnewline
5 & diffusion  & \multicolumn{2}{c}{2040 } & \multicolumn{2}{c}{0 } & \multicolumn{2}{c}{0 } & \multicolumn{2}{c}{} & 0&37  & 0&77  & 0&70  & \multicolumn{2}{c}{} & \textbf{0}&\textbf{87 } & \textbf{-0}&\textbf{08}\tabularnewline
6 & double\_perovskite\_gap  & \multicolumn{2}{c}{6529 } & \multicolumn{2}{c}{1 } & \multicolumn{2}{c}{0 } & \multicolumn{2}{c}{} & 0&48  & \textbf{0}&\textbf{82 } & 0&77  & \multicolumn{2}{c}{} & \textbf{0}&\textbf{91 } & \textbf{-0}&\textbf{1}\tabularnewline
7 & elastic\_tensor  & \multicolumn{2}{c}{5905 } & \multicolumn{2}{c}{0 } & \multicolumn{2}{c}{0 } & \multicolumn{2}{c}{} & 0&57  & \textbf{0}&\textbf{92 } & \textbf{1 }&\textbf{00} & \multicolumn{2}{c}{} & \textbf{8}&\textbf{2e2}  & \textbf{0}&\textbf{09}\tabularnewline
8 & exfoliation\_E  & \multicolumn{2}{c}{3180 } & \multicolumn{2}{c}{0 } & \multicolumn{2}{c}{0 } & \multicolumn{2}{c}{} & \textbf{0}&\textbf{92 } & \textbf{0}&\textbf{98 } & \textbf{0}&\textbf{97 } & \multicolumn{2}{c}{} & 1&14  & \textbf{-0}&\textbf{41}\tabularnewline
9 & hea\_hardness  & \multicolumn{2}{c}{1850 } & \multicolumn{2}{c}{0 } & \multicolumn{2}{c}{0 } & \multicolumn{2}{c}{} & 0&30  & \textbf{0}&\textbf{82 } & 0&73  & \multicolumn{2}{c}{} & 0&91  & -0&03\tabularnewline
10 & heusler  & \multicolumn{2}{c}{5405 } & \multicolumn{2}{c}{0 } & \multicolumn{2}{c}{0 } & \multicolumn{2}{c}{} & \textbf{0}&\textbf{66 } & \textbf{0}&\textbf{89 } & 0&79  & \multicolumn{2}{c}{} & 0&96  & \textbf{-0}&\textbf{05}\tabularnewline
11 & Li\_conductivity  & \multicolumn{2}{c}{1860 } & \multicolumn{2}{c}{0 } & \multicolumn{2}{c}{0 } & \multicolumn{2}{c}{} & \textbf{0}&\textbf{60 } & \textbf{0}&\textbf{86 } & 0&75  & \multicolumn{2}{c}{} & 0&93  & 0&00\tabularnewline
12 & metallicglass\_Dmax  & \multicolumn{2}{c}{4990 } & \multicolumn{2}{c}{0 } & \multicolumn{2}{c}{3 } & \multicolumn{2}{c}{} & \textbf{0}&\textbf{89 } & \textbf{0}&\textbf{96 } & \textbf{0}&\textbf{88 } & \multicolumn{2}{c}{} & 0&88  & \textbf{-0}&\textbf{28}\tabularnewline
13 & metallicglass\_Rc  & \multicolumn{2}{c}{8412 } & \multicolumn{2}{c}{2213 } & \multicolumn{2}{c}{7825 } & \multicolumn{2}{c}{} & \textbf{1}&\textbf{00 } & \textbf{1}&\textbf{00 } & \textbf{1}&\textbf{00 } & \multicolumn{2}{c}{} & \textbf{1}&\textbf{7e+26}  & \textbf{-0}&\textbf{27}\tabularnewline
14 & metallicglass\_Rc\_LLM  & \multicolumn{2}{c}{1485 } & \multicolumn{2}{c}{0 } & \multicolumn{2}{c}{8 } & \multicolumn{2}{c}{} & \textbf{0}&\textbf{81 } & \textbf{0}&\textbf{96 } & \textbf{0}&\textbf{94 } & \multicolumn{2}{c}{} & 0&99  & \textbf{-0}&\textbf{25}\tabularnewline
15 & Mg\_alloy  & \multicolumn{2}{c}{1825 } & \multicolumn{2}{c}{0 } & \multicolumn{2}{c}{0 } & \multicolumn{2}{c}{} & 0&42  & \textbf{0}&\textbf{86 } & \textbf{0}&\textbf{81 } & \multicolumn{2}{c}{} & \textbf{0}&\textbf{88 } & \textbf{-0}&\textbf{08}\tabularnewline
16 & oxide\_vacancy  & \multicolumn{2}{c}{24570 } & \multicolumn{2}{c}{0 } & \multicolumn{2}{c}{0 } & \multicolumn{2}{c}{} & 0&46  & \textbf{0}&\textbf{89 } & 0&79  & \multicolumn{2}{c}{} & 0&97  & \textbf{0}&\textbf{06}\tabularnewline
17 & perovskite\_ASR  & \multicolumn{2}{c}{1445 } & \multicolumn{2}{c}{0 } & \multicolumn{2}{c}{0 } & \multicolumn{2}{c}{} & 0&34  & 0&68  & 0&67  & \multicolumn{2}{c}{} & 0&97  & -0&03\tabularnewline
\hline 
\end{tabular}

\caption{\label{tab:Stats-1}Datasets statistics: $M$ (size), $n_{0}(u_{E})$
(number of non-positive uncertainties), $n_{0}(E)$ (number of null
errors), $\beta_{GM}(u_{E}^{2})$, $\beta_{GM}(E^{2})$, $\beta_{GM}(Z^{2})$,
$ZMS$, $\zeta_{ZMS}$, $RCE$ and $\zeta_{RCE}$ . For skewness statistics,
the bold data point to potentially problematic values. For calibration
statistics the bold values point to invalid calibration. }
\end{table}
\begin{table}[t]
\begin{tabular}{llr@{\extracolsep{0pt}.}lr@{\extracolsep{0pt}.}lr@{\extracolsep{0pt}.}lr@{\extracolsep{0pt}.}lr@{\extracolsep{0pt}.}lr@{\extracolsep{0pt}.}lr@{\extracolsep{0pt}.}lr@{\extracolsep{0pt}.}lr@{\extracolsep{0pt}.}lr@{\extracolsep{0pt}.}l}
\hline 
\# & Name  & \multicolumn{2}{c}{$M$} & \multicolumn{2}{c}{$n_{0}(u_{E})$ } & \multicolumn{2}{c}{$n_{0}(E)$} & \multicolumn{2}{c}{} & \multicolumn{2}{c}{$\beta_{GM}(u_{E}^{2})$} & \multicolumn{2}{c}{$\beta_{GM}(E^{2})$} & \multicolumn{2}{c}{$\beta_{GM}(Z^{2})$} & \multicolumn{2}{c}{} & \multicolumn{2}{c}{ZMS } & \multicolumn{2}{c}{RCE}\tabularnewline
\hline 
18 & perovskite\_conductivity  & \multicolumn{2}{c}{36150 } & \multicolumn{2}{c}{0 } & \multicolumn{2}{c}{0 } & \multicolumn{2}{c}{} & \textbf{0}&\textbf{69 } & \textbf{0}&\textbf{94 } & \textbf{0}&\textbf{88 } & \multicolumn{2}{c}{} & 0&99  & \textbf{0}&\textbf{05}\tabularnewline
19 & perovskite\_formationE  & \multicolumn{2}{c}{28938 } & \multicolumn{2}{c}{0 } & \multicolumn{2}{c}{1 } & \multicolumn{2}{c}{} & 0&24  & 0&67  & 0&65  & \multicolumn{2}{c}{} & 0&99  & \textbf{-0}&\textbf{01}\tabularnewline
20 & perovskite\_Habs  & \multicolumn{2}{c}{3975 } & \multicolumn{2}{c}{0 } & \multicolumn{2}{c}{0 } & \multicolumn{2}{c}{} & \textbf{0}&\textbf{85 } & \textbf{0}&\textbf{96 } & \textbf{0}&\textbf{81 } & \multicolumn{2}{c}{} & 1&1  & -0&05\tabularnewline
21 & perovskite\_Opband  & \multicolumn{2}{c}{14560 } & \multicolumn{2}{c}{0 } & \multicolumn{2}{c}{0 } & \multicolumn{2}{c}{} & \textbf{0}&\textbf{71 } & \textbf{0}&\textbf{92 } & \textbf{0}&\textbf{81 } & \multicolumn{2}{c}{} & 1&08  & \textbf{0}&\textbf{06}\tabularnewline
22 & perovskite\_stability  & \multicolumn{2}{c}{14220 } & \multicolumn{2}{c}{0 } & \multicolumn{2}{c}{0 } & \multicolumn{2}{c}{} & \textbf{0}&\textbf{65 } & \textbf{0}&\textbf{93 } & \textbf{0}&\textbf{85 } & \multicolumn{2}{c}{} & 1&07  & -0&02\tabularnewline
23 & perovskite\_tec  & \multicolumn{2}{c}{2055 } & \multicolumn{2}{c}{0 } & \multicolumn{2}{c}{0 } & \multicolumn{2}{c}{} & \textbf{0}&\textbf{62 } & \textbf{0}&\textbf{87 } & \textbf{0}&\textbf{81 } & \multicolumn{2}{c}{} & 1&01  & \textbf{-0}&\textbf{07}\tabularnewline
24 & perovskite\_workfunction  & \multicolumn{2}{c}{3065 } & \multicolumn{2}{c}{0 } & \multicolumn{2}{c}{0 } & \multicolumn{2}{c}{} & 0&42  & 0&70  & 0&63  & \multicolumn{2}{c}{} & \textbf{0}&\textbf{86 } & \textbf{-0}&\textbf{07}\tabularnewline
24 & phonon\_freq  & \multicolumn{2}{c}{6325 } & \multicolumn{2}{c}{0 } & \multicolumn{2}{c}{0 } & \multicolumn{2}{c}{} & \textbf{0}&\textbf{93 } & \textbf{0}&\textbf{95 } & \textbf{0}&\textbf{73 } & \multicolumn{2}{c}{} & 0&97  & \textbf{-0}&\textbf{06}\tabularnewline
26 & piezoelectric  & \multicolumn{2}{c}{4705 } & \multicolumn{2}{c}{0 } & \multicolumn{2}{c}{0 } & \multicolumn{2}{c}{} & 0&30  & \textbf{0}&\textbf{83 } & \textbf{0}&\textbf{82 } & \multicolumn{2}{c}{} & 1&01  & 0&00\tabularnewline
27 & RPV\_TTS  & \multicolumn{2}{c}{22675 } & \multicolumn{2}{c}{0 } & \multicolumn{2}{c}{0 } & \multicolumn{2}{c}{} & 0&58  & \textbf{0}&\textbf{83}  & 0&72  & \multicolumn{2}{c}{} & 0&98  & \textbf{-0}&\textbf{03}\tabularnewline
28 & semiconductor\_lvls  & \multicolumn{2}{c}{4480 } & \multicolumn{2}{c}{0 } & \multicolumn{2}{c}{0 } & \multicolumn{2}{c}{} & 0&41  & 0&79  & 0&74  & \multicolumn{2}{c}{} & \textbf{0}&\textbf{92 } & \textbf{-0}&\textbf{03}\tabularnewline
29 & steel\_yield  & \multicolumn{2}{c}{1560 } & \multicolumn{2}{c}{0 } & \multicolumn{2}{c}{0 } & \multicolumn{2}{c}{} & 0&62  & \textbf{0}&\textbf{88 } & 0&73  & \multicolumn{2}{c}{} & 0&92  & -0&03\tabularnewline
30 & superconductivity  & \multicolumn{2}{c}{18756 } & \multicolumn{2}{c}{0 } & \multicolumn{2}{c}{0 } & \multicolumn{2}{c}{} & 0&48  & \textbf{0}&\textbf{88 } & \textbf{0}&\textbf{8 } & \multicolumn{2}{c}{} & 1&00  & \textbf{-0}&\textbf{02}\tabularnewline
31 & thermal\_conductivity  & \multicolumn{2}{c}{13080 } & \multicolumn{2}{c}{0 } & \multicolumn{2}{c}{0 } & \multicolumn{2}{c}{} & \textbf{0}&\textbf{86 } & \textbf{0}&\textbf{93 } & \textbf{0}&\textbf{87 } & \multicolumn{2}{c}{} & 0&95  & \textbf{-0}&\textbf{12}\tabularnewline
32 & thermalcond\_aflow  & \multicolumn{2}{c}{24435 } & \multicolumn{2}{c}{0 } & \multicolumn{2}{c}{0 } & \multicolumn{2}{c}{} & \textbf{0}&\textbf{92 } & \textbf{0}&\textbf{97 } & 0&75  & \multicolumn{2}{c}{} & 0&96  & 0&02\tabularnewline
33 & thermalexp\_aflow  & \multicolumn{2}{c}{24430 } & \multicolumn{2}{c}{0 } & \multicolumn{2}{c}{0 } & \multicolumn{2}{c}{} & \textbf{0}&\textbf{87 } & \textbf{0}&\textbf{97 } & \textbf{0}&\textbf{89 } & \multicolumn{2}{c}{} & 1&17  & \textbf{-0}&\textbf{11}\tabularnewline
\hline 
\end{tabular}

\caption{\label{tab:Stats2-1}Table \ref{tab:Stats-1}, followed.}
\end{table}

\end{document}